\ifdefined\pdftexversion\pdfoutput=1\fi
\documentclass{civilcadbench}
\usepackage{booktabs}
\usepackage{multirow}
\usepackage{xspace}
\usepackage{amssymb}
\usepackage{pifont}
\newcommand{\cmark}{\ding{51}}
\newcommand{\xmark}{\ding{55}}
\newcommand{\pmark}{$\circ$}
\newcommand{\ours}{CivilCADBench\xspace}
\newcommand{\eg}{\emph{e.g.}\xspace}

\title{CivilCADBench: Benchmarking Multimodal LLMs for\\
Understanding, Reconstructing, and Designing\\
Civil Engineering Drawing Sets}

\author{Zelin Zheng$^{1}$ \quad Difei Wu$^{2,\dagger}$ \quad Xinyan Liu$^{1,3}$ \quad Minghong Chen$^{2}$\\
Antoni B. Chan$^{3}$ \quad Difan Zou$^{4,\dagger}$}
\affiliation{$^{1}$Harbin Institute of Technology (Weihai) \quad $^{2}$Tongji University\\
$^{3}$City University of Hong Kong \quad $^{4}$The University of Hong Kong \quad $^{\dagger}$Corresponding authors}
\projectpage{https://agenticivil.com/civilcadbench/}
\date{}

\begin{document}
\maketitle
\begin{abstract}
Civil infrastructure is designed, built, and maintained through engineering drawing sets, yet preparing and revising them remains labour-intensive and calls for automation. Multimodal large language models (MLLMs) are promising candidates, but assessing them requires a benchmark grounded in real engineering practice. We introduce \ours, a benchmark for understanding, reconstructing, and designing civil engineering drawing sets, built from real infrastructure drawings and source CAD files containing over 250{,}000 CAD entities. Unlike parametric 3D CAD models, these drawings are dense, layered, and multi-sheet; and since prints are lossy projections with no pixel trace of layers, a file can print correctly yet remain wrongly organised.  \ours comprises three task families of increasing difficulty: understanding tasks are evaluated with 2{,}183 expert-reviewed questions ranging from single sheets to whole sets, while reconstruction and design tasks are agentic, requiring models to rebuild printed pages as editable CAD files, either as single pages or entire drawing sets, and to produce new designs. These tasks are performed across 232 pages, 28 sets, and 19 design tasks, and evaluated  by a deterministic scorer that compares model outputs with the designers' source files on geometry, layers, text, and dimensions. We evaluate seven frontier MLLMs on \ours, and find that they read and redraw printed content accurately but recover the file structure poorly: the best model scores above 90\% in understanding and single-page reconstruction, yet only 60\% on layer assignment and 48\% on design. Grounded in real engineering practice, \ours exposes both the capabilities and the limitations of frontier models in professional civil engineering, laying a foundation for next-generation models and agentic systems for intelligent civil engineering.
\end{abstract}

\begin{center}
\centering
\includegraphics[width=\linewidth]{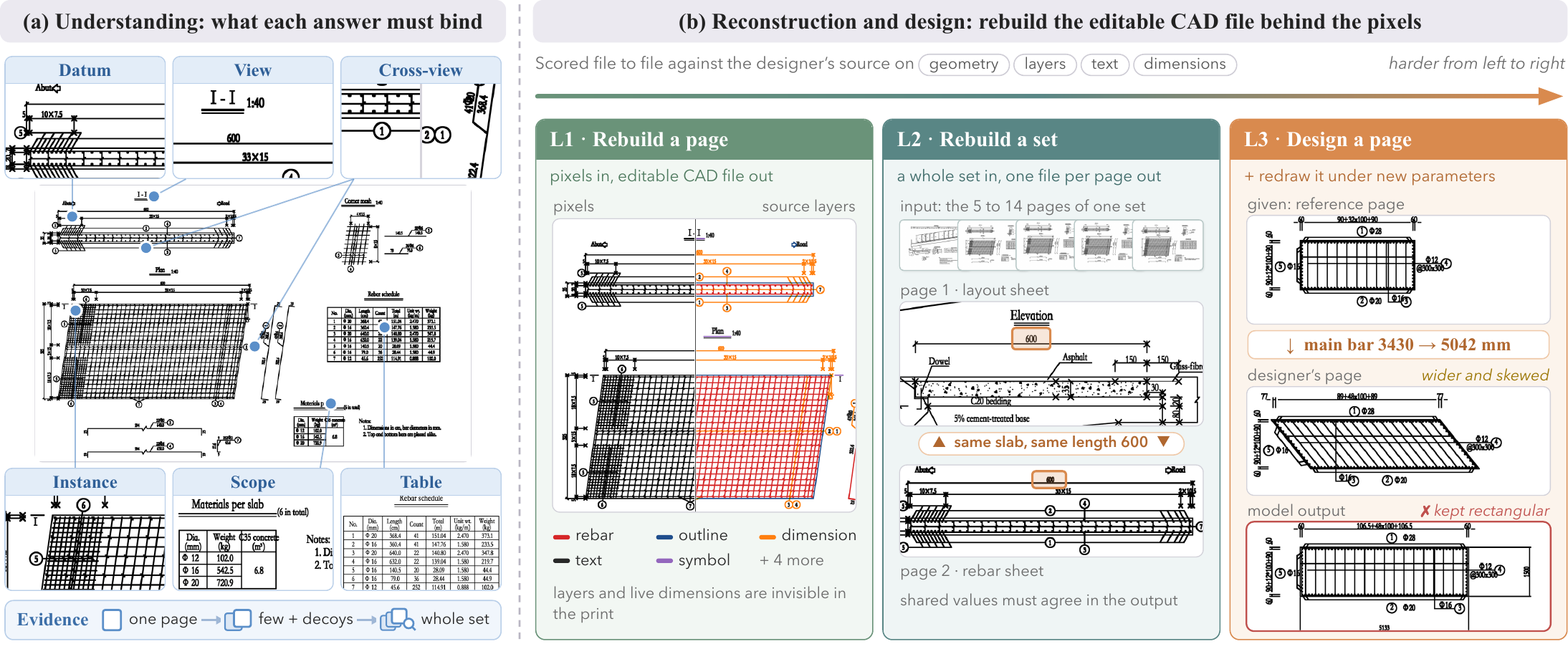}
\captionof{figure}{\textbf{Overview of \ours.} \ours covers the drawing work of civil engineering: understanding a drawing, reconstructing the CAD file behind it and designing a new sheet.
(a) Understanding: multi-select questions are written from binding points, where a printed value must be tied to the object it describes (six types shown), and are asked over one page, a few pages or a whole unit.
(b) Reconstruction and design, in rising difficulty: rebuild one page as an editable CAD file (L1), a whole set whose shared values must agree (L2) and a withheld page from its reference page and new parameters (L3), all scored against the source file.}
\label{fig:teaser}
\end{center}

\section{Introduction}
\label{sec:intro}

Civil infrastructure depends on drawing sets throughout its life. A set contains tens to hundreds of sheets and fixes every dimension and quantity in the project, down to individual reinforcing bars~\cite{structdrawagent}. The same set accompanies the project from design through construction to long-term maintenance, yet each stage and each party uses it differently. Reviewers and contractors \textit{understand} the printed sheets, and most parties hold prints rather than source files~\cite{paperdrawings}. The designer keeps the CAD files, and a print is a lossy projection of them: the layer assignment of each entity disappears. This loss becomes a practical problem when an existing structure must be maintained or repaired. Engineers often have only the record drawings~\cite{akanbi2022,zhao2021bim,zhang2021bimreview}, so they must \textit{reconstruct} an editable file from those prints. Reconstruction has to recover the structure that printing discarded while keeping shared values consistent across sheets. Designers face a similar constraint when they \textit{design} new sheets by adapting existing files to new parameters, because the result must remain coherent with the rest of the set. This work remains largely manual and repetitive~\cite{drafterbench}, yet it requires a correct reading of the drawing at every step, which makes it a natural target for multimodal large language models (MLLMs) acting as agents.

Research on CAD generation has so far targeted parametric 3D models: methods and benchmarks generate construction sequences or CAD programs for solids from text, images, point clouds or three-view drawings~\cite{deepcad,text2cad,cadrecode,benchcad,cadbench,paramcadv3,llmcadsurvey}, and success is measured mainly on the resulting shape. However, civil engineering drawings are dense, layered 2D sheets delivered as sets rather than parametric solids, so a usable file must preserve layer semantics and cross-sheet consistency, neither of which a shape metric captures. One sheet holds hundreds to thousands of entities, and every entity lies on a named layer such as structure outline, reinforcement or dimensions~\cite{iso13567}. Layers let an engineer select, hide or edit all entities of one kind at once, but they leave no trace in the pixels of a print, so moving every reinforcement bar onto the outline layer yields the same pixels but a worse file (\cref{fig:samepix}). The same member also recurs across plans, sections and schedules, so its values must be read and written consistently over many sheets.

In the civil engineering domain, existing evaluations have largely centred on understanding. Drawing benchmarks ask questions about construction, architectural and large-format sheets~\cite{drawingvqa,aecvbench,ledbench}, among which AEC-Bench~\cite{aecbench} comes closest to practice: it has agents review multi-sheet drawing sets and specifications for injected defects. These benchmarks test how well a model reads what is printed and leave open whether it can deliver the file that reconstruction and design depend on. Only a few works go beyond understanding, and even they evaluate something other than the drawing file itself: DrafterBench~\cite{drafterbench} grades the tool calls that revise a drawing, and TriView2CAD~\cite{creftcad} reads fifteen parameters from three-view drawings of bridge piers. A benchmark close to engineering practice is therefore needed to measure whether MLLMs can actually carry out drawing work against the real artifacts and design source files engineers rely on.

We introduce \ours (\cref{fig:teaser}), a benchmark for understanding, reconstructing and designing civil engineering drawing sets, built from the source CAD files of real infrastructure projects.
Its pages are complete construction sheets: the source files comprise over 250{,}000 CAD entities, with a median of 523 per page.
In \textit{understanding}, a model answers 1{,}800 single-sheet and 383 multi-sheet questions; all were reviewed by civil engineering experts, and the multi-sheet ones are paired with short-answer probes that locate what a model misreads.
\textit{Reconstruction} and \textit{design} form three levels of rising difficulty: an agent rebuilds one printed page as an editable CAD file (\textit{L1}, 232 pages), rebuilds a whole set whose shared values must agree across pages (\textit{L2}, 28 sets), and designs a withheld page from the page it was derived from and the new parameters, as designers do when a member changes size (\textit{L3}, 19 tasks).

We evaluate seven MLLMs and find a consistent pattern: models read and redraw what a drawing shows well, but recover the structure of the file behind it poorly.
The best model scores over 93\% in understanding and 91\% in single-page reconstruction, yet only 60\% on layer assignment, which leaves no trace in the pixels, and 48\% on design, where values must be computed.
Open models come close only in single-page reconstruction (81\%) and trail in understanding (64\%), set reconstruction and design.
Since every ground truth is paired with its print and scoring is deterministic, \ours measures the abilities that these models lack and provides label-free data to train them.
In summary, this paper makes three contributions.
\begin{itemize}
\item We introduce \ours, to our knowledge the first benchmark to evaluate the understanding, reconstruction and design of engineering drawings together, on the drawing sets of real infrastructure projects with the designers' layered source files as ground truth.
\item We build the data and the scorer so that no manual labels are needed: each input page and its ground truth come from one source file, and a deterministic scorer, verified by fault injection, compares files on geometry, layers, text and dimensions, including what the print hides.
\item We show that frontier MLLMs reproduce what a drawing shows but fall well short in organising it into layers and in designing a new page, and our analysis traces this limit to the structure behind the drawing, namely its layer semantics and value dependencies, while reading a drawing well turns out to predict little about producing it.
\end{itemize}

\begin{table*}[t]
\centering
\caption{\textbf{Comparison of \ours with the closest benchmarks.} U, R, D: understanding, reconstruction and design tasks. Real: data from real projects rather than authored or synthetic. Sets: tasks need several sheets of one set. Layers: named layers are part of the target. Det.: scoring is deterministic, with no LLM or human judge. \pmark: partly; --: not stated.}
\label{tab:compare}
\footnotesize
\setlength{\tabcolsep}{6pt}
\begin{tabular}{@{}ll ccc cccc l@{}}
\toprule
Benchmark & Domain & U & R & D & Real & Sets & Layers & Det. & Scored on \\
\midrule
\multicolumn{10}{@{}l}{\textit{3D CAD models}} \\
CADBench~\cite{cadbench} & mechanical parts & \xmark & \cmark & \xmark & \pmark & \xmark & \xmark & \cmark & 3D shape \\
BenchCAD~\cite{benchcad} & mechanical parts & \cmark & \cmark & \cmark & \xmark & \xmark & \xmark & \cmark & 3D shape \\
Param.\ CAD Bench~\cite{paramcadv3} & mechanical parts & \xmark & \cmark & \cmark & \xmark & \xmark & \xmark & \cmark & 3D shape \\
\midrule
\multicolumn{10}{@{}l}{\textit{Engineering drawings}} \\
MechVQA~\cite{mechvqa} & mechanical drawings & \cmark & \xmark & \xmark & \cmark & \xmark & \xmark & \xmark & answers \\
LED-BENCH~\cite{ledbench} & large-format sheets & \cmark & \xmark & \xmark & \cmark & \xmark & \xmark & \xmark & answers \\
DrawingVQA~\cite{drawingvqa} & building structures & \cmark & \xmark & \xmark & \cmark & \pmark & \xmark & \cmark & answers \\
AEC-Bench~\cite{aecbench} & building projects & \cmark & \xmark & \xmark & \cmark & \cmark & \xmark & \cmark & findings \\
TriView2CAD~\cite{creftcad} & bridge piers & \cmark & \xmark & \xmark & \pmark & \xmark & \xmark & \cmark & parameters \\
DrafterBench~\cite{drafterbench} & civil drawings & \xmark & \xmark & \pmark & \cmark & \xmark & \xmark & \cmark & tool calls \\
AutoCAD Bench~\cite{autocadbench} & 2D and 3D CAD & \xmark & \cmark & \xmark & -- & \xmark & -- & \pmark & gold file \\
\midrule
\textbf{\ours} & civil drawing sets & \cmark & \cmark & \cmark & \cmark & \cmark & \cmark & \cmark & answers, source file \\
\bottomrule
\end{tabular}
\end{table*}

\section{Related Work}
\label{sec:related}

\noindent\textbf{CAD generation benchmarks.}
Our task is closest to CAD generation, where models write CAD through code, with methods that produce construction sequences or programs for 3D models from text, point clouds or drawings~\cite{deepcad,text2cad,cadrecode,drawing2cad}, and recent benchmarks evaluate CAD programs written by language models and agents~\cite{benchcad,cadbench,paramcadv3,cadengbench,cadworld,engiworld,gptastracad}.
They measure mainly the resulting shape, whereas a 2D drawing set also carries meaning in layers, text, dimensions and agreement between sheets, which call for file-level scoring.
We therefore keep the protocol of writing CAD through code and score the file itself.

\noindent\textbf{Engineering drawing benchmarks.}
Drawings have mostly been studied as something to read, from symbol spotting on floor plans~\cite{floorplancad,archcad400k} to question answering with MLLMs on mechanical drawings~\cite{mechvqa,mechreason}, construction drawings~\cite{drawingvqa}, architectural plans~\cite{aecvbench,mmarch,archsibench} and large-format sheets~\cite{ledbench}, and to coding agents that review multi-sheet building projects~\cite{aecbench}.
They measure how well a model reads a drawing and leave open whether it can deliver one; the few works that go further target proxies of the file, such as tool calls~\cite{drafterbench}, bridge-pier parameters~\cite{creftcad}, floor plans~\cite{housemind} or redrawn drawings~\cite{autocadbench}, and none scores a whole set against the designer's layered source file (\cref{tab:compare}; \cref{sec:s-related}).
Our understanding track borrows from long-document benchmarks~\cite{mpdocvqa,slidevqa,mmlongbenchdoc,docscope} the practice of giving each question's evidence pages and asking over the whole unit, and adds short-answer probes~\cite{press2023,wemath} that show where a wrong answer comes from.

\noindent\textbf{From raster images to editable files.}
Vectorisation recovers primitives from technical drawings and floor plans~\cite{egiazarian2020,floorplanvlm}, and with MLLMs image-to-structure tasks have spread to charts, webpages, slides and vector graphics~\cite{chartmimic,design2code,image2struct,pptbench,starvector,svgenius,drawai}.
These works compare renderings or ask a model to judge, and layer membership stays invisible to both (\cref{fig:samepix}); in order to make it visible, we score the file against its source.

\section{\ours}
\label{sec:benchmark}

\begin{figure*}[t]
\centering
\includegraphics[width=\linewidth]{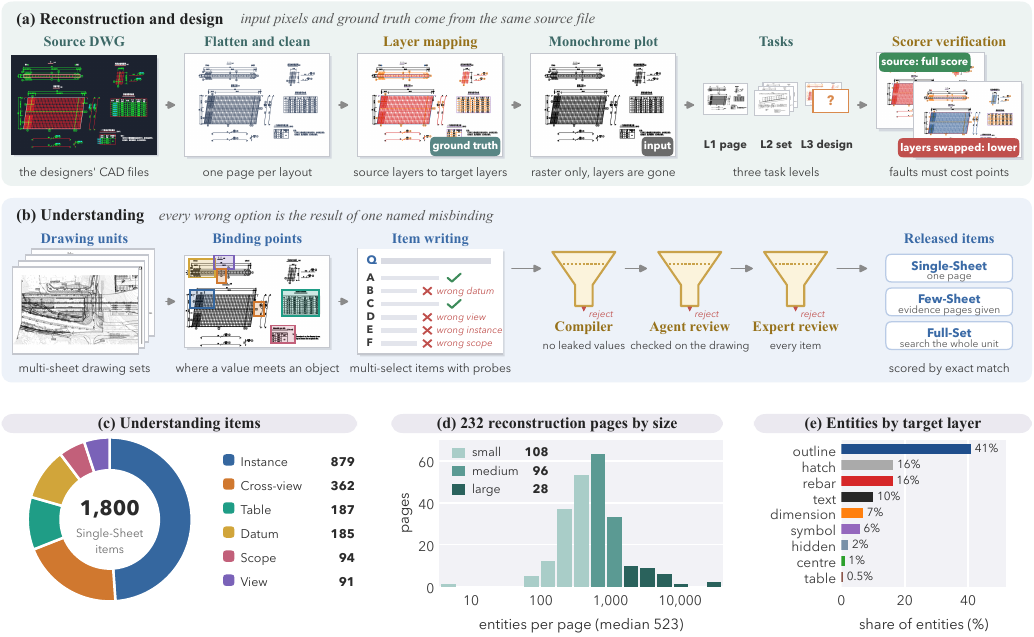}
\caption{\textbf{Construction and contents of \ours.}
(a) Reconstruction and design: each layout of the designers' CAD files is flattened into one page, its layers are mapped to nine target layers to give the ground truth, and a monochrome plot of the same file is the input; faults injected into the ground truths verify the scorer.
(b) Understanding: items are written from binding points on real drawing sets, and a compiler, an independent agent and civil engineering experts check every item. Steps in amber are checks.
(c) Single-Sheet items by binding type. (d) Entities per page of the reconstruction pages, by size tier. (e) Share of entities on each target layer.}
\label{fig:data}
\end{figure*}

\begin{table}[t]
\centering
\caption{\textbf{Statistics of \ours.}}
\label{tab:stats}
\footnotesize
\setlength{\tabcolsep}{3pt}
\begin{tabular}{@{}llr@{}}
\toprule
Track & Unit & Count \\
\midrule
\multirow{4}{*}{\shortstack[l]{Reconstruction\\and design}} & L1 pages & 232 \\
 & Entities (median per page) & 257.7k (523) \\
 & L2 sets (pages) & 28 (230) \\
 & L3 design tasks & 19 \\
\midrule
\multirow{3}{*}{Understanding} & Single-Sheet items & 1{,}800 \\
 & Few-Sheet parents / probes & 383 / 1{,}953 \\
 & Full-Set parents / probes & 330 / 1{,}989 \\
\bottomrule
\end{tabular}
\end{table}

In order to measure drawing work as engineers actually do it, we build \ours on three principles.
First, the target is the designers' own file, and the input is a plot of that same file, so no annotation is needed.
Second, we score files, because printing discards the layers and only a comparison of files can check them.
Third, our questions target binding, since the difficulty of reading a drawing lies in tying each number to its object.

\subsection{Task formulation}
\label{sec:formulation}

To make precise what printing throws away, we model a page as a file $F$, a set of entities that each have a geometry, a type (line, arc, hatch, text or dimension), a layer and, for text and dimensions, a value, and a drawing set as $D=\{F_1,\dots,F_n\}$ whose pages share values that must agree.
The designer draws each page from parameters $\theta$, such as a slab width, under the rules of the design office, $F=G(\theta)$. The plotting $\pi$ maps a file to pixels and is many-to-one: it drops the layers, turns a dimension into lines and text, and cuts the link between the printed values and $\theta$.

Each task family then starts from the print and asks for one more thing that plotting left out (\cref{tab:stats}).
\textit{Understanding} asks only for what is printed, $a=U(\pi(D),q)$; its three tiers differ in how much of $D$ must be searched.
\textit{Reconstruction} asks for the structure plotting dropped: the model inverts the plot, $\hat{D}=R(\pi(D))$, for one page (L1) or a set (L2), where shared values must also agree, and because $\pi$ is many-to-one we compare $\hat{D}$ with $D$ itself.
\textit{Design} asks for the rules behind the values: given a reference page $F$, the prints of the rest of the set and a change from $\theta$ to $\theta'$, produce $F'=G(\theta')$ (L3); \cref{sec:exp,sec:analysis} follow this ladder.

\subsection{Benchmark construction}
\label{sec:gt}

Construction drawings arrive as coordinated sets in which the same member recurs across plans, sections, and schedules, so a model that reads only a print must recover editable geometry together with the cross-page consistency that printing discards. We build the reconstruction data from 232 pages of bridge and tunnel construction drawings and their source DWG files (>250{,}000 entities; median 523 per page), flatten each paper-space layout into one A3 page, and save it as a DXF file, an open text format that code can read and write, to serve as ground truth (\cref{fig:data}a; \cref{sec:s-data}). The input is a monochrome office-style plot of the same file rasterised at 200\,dpi into a PDF with no vector or text layer, so the agent sees exactly $\pi(F)$ while the scorer sees $F$. Because line weight and line type are stored on each entity rather than on its layer, the print preserves what an entity is for but not the layer that names it. Since offices name layers freely, we map every source layer through per-project rules to one of nine functional layers (structure outline, reinforcement, hidden lines, centre lines, hatch, dimensions, text, symbols, and tables) and give the agent the same nine names, which requires it to infer each entity's role from what the entity depicts. We also split pages into three size tiers by entity count (\cref{fig:data}d) to verify that page density alone does not drive the results.

\noindent\textbf{Reconstruction and design tasks.}
\label{sec:rebuild}
These tasks mirror practice at increasing scope and require the agent to produce editable drawings rather than descriptions. \textit{L1} places the agent as an engineer holding one record drawing and provides \texttt{input.pdf} plus a task statement that fixes the coordinate convention, the nine layers, and the expected entities, after which the agent writes \texttt{output.dxf} in a sandbox with Python, \texttt{ezdxf}~\cite{ezdxf}, and a preview tool (\cref{sec:s-episode}). \textit{L2} asks for a whole set because projects are delivered as sets and the same member recurs across plans, sections, and schedules; the agent receives all pages of one set (5 to 14), formed from the drawing numbers of one project, and writes one DXF per page with consistent layers and styles and identical values for recurring members. \textit{L3} follows how designers derive some pages from a sibling page of the same drawing number with new parameters, such as approach slabs of two widths, so the derived page is withheld as ground truth while the agent receives the sibling's DXF and raster page, the rest of the set as rasters, and the new parameters (\eg the slab widens from 325 to 350\,cm with all else unchanged). We construct each task so that every changed value traces to the given parameters or pages, and we remove unreferenced blocks from the reference DXF, which ensures that parameter-dependent quantities such as bar counts, lengths, and schedule rows must change while everything else stays; we identify those changes by comparing the withheld page with its sibling.

\noindent\textbf{Understanding tasks.}
\label{sec:read}
Binding is the central thread of these tasks because a printed value matters only once it is tied to the object it describes, so we build items around binding points (\cref{fig:data}b) such as a dimension line's endpoints, an elevation's datum, a section mark, a numbered schedule, a counting scope, or a cross-view reference. Each wrong option instantiates a specific misbinding that the drawing itself refutes, which makes a wrong choice diagnostic of the binding that failed. We design stems to describe the task without naming the trap and allow at most two options to concern the same attribute, so the wording never reveals the answer. \textit{Single-Sheet} items are answerable from one page; \textit{Few-Sheet} parents span several pages and include their evidence pages plus two look-alike decoys (same-format sheets from a different instance or competing tables); and \textit{Full-Set} poses the same parents over the whole $\sim$30-page unit, where the model acts as an agent with a \texttt{view\_page} tool for whole pages or zoomed boxes and a calculator within 16 steps and 24 page views, so comparing the last two tiers separates reading the evidence from finding it. Because a multi-sheet answer can fail by misreading a fact or by failing to combine facts, each parent includes short-answer probes that together determine every option. \textit{Read} probes name the object and ask for a value, whereas \textit{bind} probes refer to the object only through a property printed on another sheet, and typed scorers check a number with its unit, named slots, or endpoint pairs. A compiler rejects stems that leak a correct-option value, answer letters are balanced so that each letter accounts for roughly one sixth of the correct options, and an LLM agent drafts each item under a written protocol while an independent second LLM agent checks it against the drawing and three civil engineering experts review every item (\cref{sec:s-experts}).
\subsection{Evaluation metrics}
\label{sec:scoring}

\begin{figure*}[t]
\centering
\includegraphics[width=\linewidth]{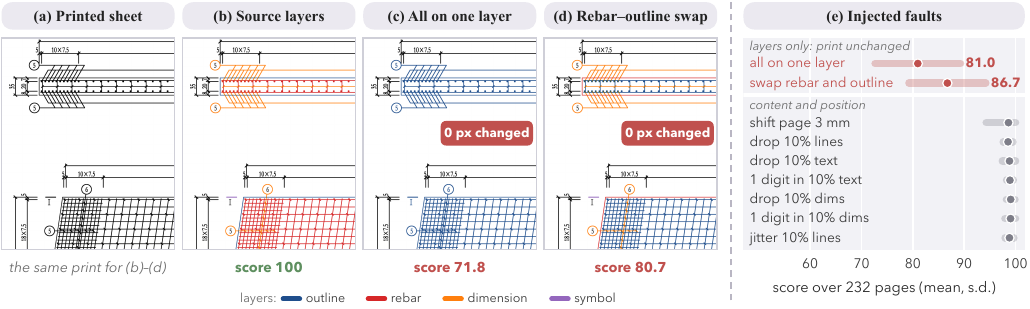}
\caption{\textbf{Three files with the same print, and fault injection.} (a) A ground-truth crop as printed. (b)--(d) Three files that all print as (a), coloured by layer: the source file (score 100), all entities on one layer (71.8) and the rebar and outline layers swapped (80.7). (e) Mean score and s.d.\ over the 232 ground truths after each injected fault; layer-only faults (red) cost the most, and no pixel metric sees them.}
\label{fig:samepix}
\end{figure*}

To keep scores reproducible, we use deterministic metrics throughout and no LLM or VLM judge.

\noindent\textbf{Reconstruction and design.}
For reconstruction and design, the scorer samples every curve at 0.25\,mm and aligns the output to the ground truth by scale and translation, so files are compared as point sets regardless of entity type. Geometry F1 $s_g$ matches points within 1\,mm; layer-aware F1 $s_l$ also requires the same target layer; text and dimension F1, $s_t$ and $s_d$, match strings within 6\,mm with similarity at least 0.8 and identical numbers, where a text may match a dimension label and vice versa because some source pages store dimensions as exploded lines and text; $s_a$ is the share of entities on the nine layers, and a missing or unreadable output scores 0. The page score is the weighted mean $S=\sum_{k\in\mathcal{C}} w_k s_k/\sum_{k\in\mathcal{C}} w_k$ over the components $\mathcal{C}$ present in the ground truth, with $(w_g,w_l,w_a,w_t,w_d)=(0.25,0.30,0.10,0.20,0.15)$. The two layer components carry the largest share (0.40), since layers turn a picture into an editable file. Two further measures target the harder levels: L2 adds \textit{shared-value recall}, the share of source values repeated across pages that the output also repeats, and L3 is scored by \textit{design recall}, the share of values changed by the new parameters that the output writes correctly. 

\noindent\textbf{Verification by fault injection.}
Before using the scorer, we check that it penalises real errors and ignores harmless differences: we inject one of ten kinds of faults into each of the 232 ground truths and score it (\cref{fig:samepix}e).
Every content fault, such as dropped text or a changed digit, lowers the score, while a 3\,mm shift of the whole page is largely absorbed by registration (98.6).
The two faults that only move entities between layers cost the most (81.0 and 86.7 on average) although every printed pixel stays the same.

\noindent\textbf{Understanding.}
Each item has six options, one to four of them correct, and is scored by exact match, so a random guess succeeds with probability 1/56. In order to separate reading from combining, we group each parent with its probes; a solution path is a set of probes that suffices to decide the parent.
\textit{Local coverage} (LC) is the share of groups where all probes of some solution path are right (the model read the facts it needs), and \textit{compositional accuracy} (CA) the share of those groups whose parent is also right (it then combined them)~\cite{press2023}.

\section{Experiments}
\label{sec:exp}
\begin{table*}[t]
\centering
\caption{\textbf{Main results on reconstruction (L1, L2), design (L3) and understanding.} \emph{L1}: page score (Score), geometry, text and dimension F1 (Geom., Text, Dim.) and layer-aware F1 (Layer). \emph{L2}: page score, its difference from L1 (vs.\ alone) and the share of values printed on several pages of a set that the output reproduces (Shared). \emph{L3}: share of changed values written correctly (Recall) and page score minus that of copying the reference page (vs.\ copy). \emph{Understanding}: exact-match accuracy per tier. $\dagger$: its API encodes an image in about 1{,}024 tokens, so each page is sent with its $2{\times}2$ tiles. Best per column in bold.}
\label{tab:main}
\footnotesize
\setlength{\tabcolsep}{3.1pt}
\begin{tabular}{@{}l ccccc ccc cc ccc@{}}
\toprule
 & \multicolumn{5}{c}{L1: rebuild a page} & \multicolumn{3}{c}{L2: rebuild a set} & \multicolumn{2}{c}{L3: design a page} & \multicolumn{3}{c@{}}{Understanding} \\
\cmidrule(lr){2-6}\cmidrule(lr){7-9}\cmidrule(lr){10-11}\cmidrule(l){12-14}
Model & Score & Geom. & Text & Dim. & Layer & Score & vs.\ alone & Shared & Recall & vs.\ copy & Single & Few & Full \\
\midrule
GPT-6~Astra~\cite{gpt6astra} & \textbf{91.2} & \textbf{95.5} & \textbf{91.7} & \textbf{94.6} & 59.9 & \textbf{87.8} & $\mathbf{-3.4}$ & \textbf{85.3} & 47.9 & $-6.0$ & \textbf{93.3} & \textbf{95.3} & \textbf{94.5} \\
GPT-6.1~Sol~\cite{gpt61sol} & 90.4 & 94.8 & 90.9 & 92.7 & 58.9 & 85.9 & $-4.5$ & 83.4 & \textbf{49.1} & $-8.9$ & 91.5 & 92.2 & 89.7 \\
GPT-5.6~Sol~\cite{gpt56sol} & 59.8 & 66.0 & 48.7 & 44.7 & 40.8 & 36.6 & $-23.2$ & 31.7 & 40.3 & $+1.0$ & 72.2 & 75.5 & 69.4 \\
\midrule
DeepSeek-V4.1-Flash$^\dagger$~\cite{deepseekv41flash} & 81.3 & 83.7 & 80.5 & 80.9 & 63.2 & 55.4 & $-25.9$ & 62.1 & 37.8 & $-2.1$ & 60.4 & 63.8 & 58.7 \\
Qwen3.8-27B~\cite{qwen38} & 36.2 & 40.5 & 38.4 & 43.2 & 30.1 & 14.5 & $-21.7$ & 18.4 & 6.3 & $-48.3$ & 58.2 & 61.4 & 53.6 \\
GLM-5.3-Flash~\cite{glm53flash} & 76.1 & 82.4 & 74.3 & 75.6 & \textbf{70.8} & 47.2 & $-28.9$ & 53.5 & 19.2 & $-4.6$ & 57.4 & 60.2 & 52.8 \\
Qwen3.5-9B~\cite{qwen35} & 5.6 & 10.4 & 6.3 & 2.5 & 1.8 & 0.0 & $-5.6$ & 0.0 & 0.0 & $-64.1$ & 34.3 & 43.1 & 8.2 \\
\bottomrule
\end{tabular}
\end{table*}

We evaluate seven MLLMs: three frontier closed models: GPT-6~Astra~\cite{gpt6astra}, GPT-6.1~Sol~\cite{gpt61sol}, and GPT-5.6~Sol~\cite{gpt56sol}; and four open models: Qwen3.8-27B~\cite{qwen38}, Qwen3.5-9B~\cite{qwen35}, DeepSeek-V4.1-Flash~\cite{deepseekv41flash} and GLM-5.3-Flash~\cite{glm53flash}. All seven are evaluated on understanding, reconstruction and design, with high reasoning effort used for all models. Reconstruction and design runs are agentic: each model writes and runs its own code in the sandbox, previews its DXF against the input, and submits once (\cref{fig:episode}), whereas Single-Sheet items and Few-Sheet parents are answered in one turn. We use Wilcoxon signed-rank tests and bootstrap intervals, plus McNemar’s test for access-tier comparisons. Our goal is to identify where model performance breaks down as the tasks increasingly require information that is discarded during printing; the results are then presented in \cref{tab:main} in the same order.

\subsection{Results on reconstruction and design tasks}
\label{sec:exp-rebuild}

We begin with single pages (L1), which test whether a model can trace a drawing.
The two GPT-6 models achieve mean scores of 91.2 and 90.4, while GPT-5.6~Sol reaches 59.8 and three of the open models 36.2 to 81.3 (\cref{tab:main}); the fourth, Qwen3.5-9B, scores 5.6 here and zero on sets and design, and the ranges below leave it out.

The picture changes once a page has to be organised as well as drawn.
Layer-aware F1 is only 59.9 and 58.9 for the GPT-6 models, and the best layer score comes from GLM-5.3-Flash (70.8), which keeps 86\% of its geometry F1 once the layer is required, against 63\% for GPT-6~Astra, so layer recovery is a separate ability from tracing (\cref{sec:analysis}).
Rebuilding a whole set (L2) is harder still: a page scores only 3.4 and 4.5 points lower inside a set than alone for the GPT-6 models but 22 to 29 points lower for the other models, and since the pages are the same, the loss comes from handling many of them in one run; shared values suffer too, with 85 and 83\% reproduced by the GPT-6 models and 18 to 62\% by the others.

\begin{figure*}[t]
\centering
\includegraphics[width=\linewidth]{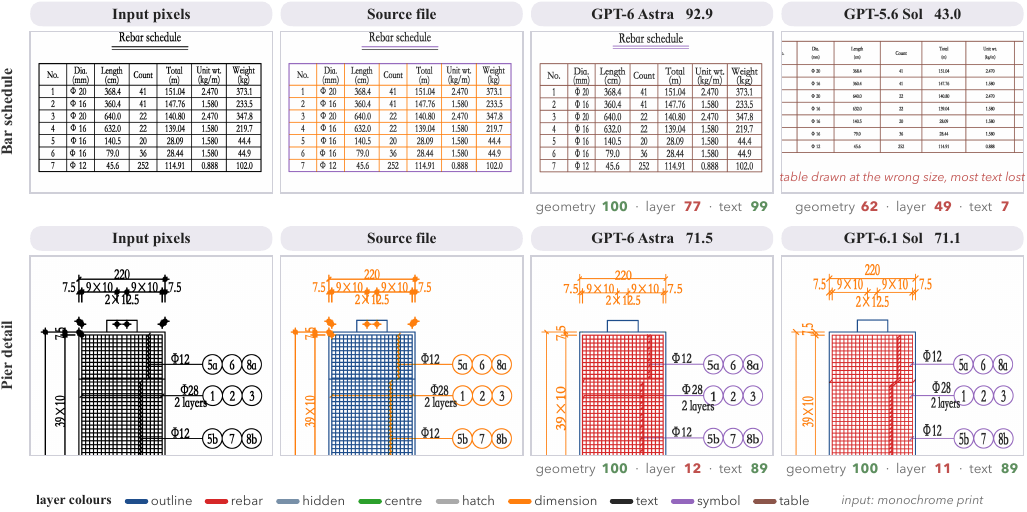}
\caption{\textbf{Qualitative comparison of single-page reconstruction (L1).} The same crop is shown in every column; the page score follows the model name, and component F1 below it is green at 85 or above and red below.
Top: GPT-6~Astra reproduces the bar schedule exactly; GPT-5.6~Sol draws the table at the wrong size and loses most text.
Bottom: both GPT-6 models redraw the pier detail line for line (geometry F1 100), but few of their lines carry the layer of the source file (layer-aware F1 12 and 11).}
\label{fig:qual}
\end{figure*}

Design (L3) is where every model falls furthest: since values must be computed, we ask how many of the changed values a model writes correctly, and even the models that trace a page accurately get fewer than half right (48 and 49\% for the GPT-6 models, 6 to 40\% for the others).
The edits also cost more than they gain: copying the reference page unchanged would score 6.0 and 8.9 points higher than the two GPT-6 models.
GPT-5.6~Sol stays level with the copy (+1.0) only because it leaves more values unchanged (\cref{sec:s-l3}), so the page score alone would favour the model that edits least.

\begin{figure}[t]
\centering
\includegraphics[width=.6\linewidth]{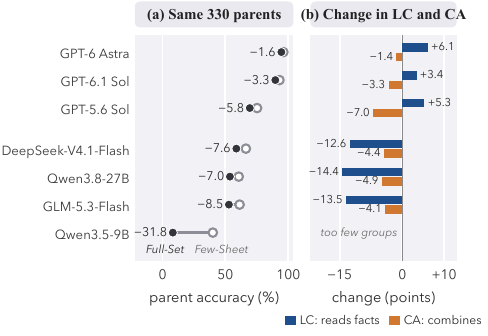}
\caption{\textbf{Understanding with the evidence given (Few-Sheet) and to be found (Full-Set).}
(a) Accuracy on the same 330 parents in both tiers.
(b) Change from Few-Sheet to Full-Set in local coverage LC (groups with all probes of a solution path right) and compositional accuracy CA (parent right given LC): finding the pages costs open models, mainly in LC, but not the strongest one.}
\label{fig:read}
\end{figure}

\begin{table}[t]
\centering
\caption{\textbf{Single-Sheet accuracy (\%) by binding type.} Inst.: instance; Cross: cross-view. $\dagger$: each page is sent with its $2{\times}2$ tiles.}
\label{tab:binding}
\footnotesize
\setlength{\tabcolsep}{2.6pt}
\begin{tabular}{@{}l cccccc@{}}
\toprule
Model & Table & Inst. & Cross & Datum & View & Scope \\
\midrule
GPT-6~Astra~\cite{gpt6astra} & 95.7 & \textbf{94.2} & \textbf{92.5} & \textbf{94.5} & \textbf{87.8} & \textbf{87.0} \\
GPT-6.1~Sol~\cite{gpt61sol} & \textbf{98.4} & 91.9 & 89.7 & 91.3 & 85.7 & 85.9 \\
GPT-5.6~Sol~\cite{gpt56sol} & 87.2 & 71.7 & 67.4 & 71.9 & 76.9 & 62.8 \\
\midrule
DeepSeek-V4.1-Flash$^\dagger$~\cite{deepseekv41flash} & 63.5 & 63.1 & 57.6 & 58.4 & 50.6 & 53.4 \\
Qwen3.8-27B~\cite{qwen38} & 75.4 & 60.2 & 51.9 & 48.1 & 57.1 & 52.1 \\
GLM-5.3-Flash~\cite{glm53flash} & 61.9 & 58.7 & 55.4 & 56.2 & 48.0 & 55.4 \\
Qwen3.5-9B~\cite{qwen35} & 55.6 & 34.2 & 31.5 & 23.8 & 24.2 & 34.0 \\
\bottomrule
\end{tabular}
\end{table}

\begin{figure*}[t]
\centering
\includegraphics[width=\linewidth]{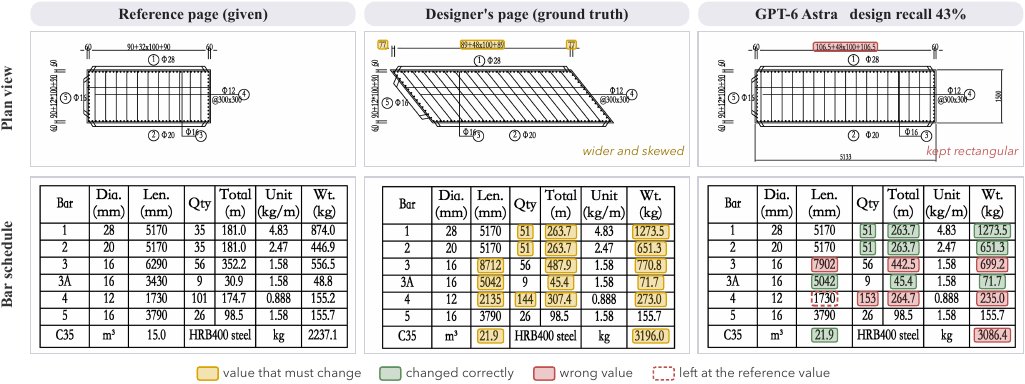}
\caption{\textbf{An example of design (L3).} The task lengthens the transverse main bar of an abutment cap (bar 3A) from 3430 to 5042\,mm. The designer's cap is wider and skewed, so counts, lengths and totals change (yellow). GPT-6~Astra gets bars 1, 2 and 3A right (green), but miscomputes bar 3, leaves bar 4 unchanged and gets the totals wrong (red): it finds the change but miscomputes it.}
\label{fig:design}
\end{figure*}

\subsection{Results on understanding tasks}
\label{sec:exp-read}

Turning to reading, with the evidence in front of them the GPT-6 models answer 89.7 to 95.3\% of the questions on every tier, close to the ceiling, while the open models answer 34 to 64\% (\cref{tab:main}).
Splitting Single-Sheet accuracy by binding type shows what separates good readers from weak ones (\cref{tab:binding}): table lookups, which resemble reading a document, are the easiest for every model, whereas datum items fall to 48.1\% for Qwen3.8-27B and 23.8\% for Qwen3.5-9B.

To test whether finding the evidence costs the open models more, we compare Few-Sheet and Full-Set on the 330 parents asked in both tiers (\cref{fig:read}a): GPT-6~Astra hardly moves (96.1 to 94.5\%), whereas GPT-6.1~Sol, GPT-5.6~Sol and Qwen3.8-27B each lose three to seven points, GLM-5.3-Flash and tiled DeepSeek-V4.1-Flash about 8, and Qwen3.5-9B from 40.0 to 8.2\%.
The probes locate the loss in reading: for Qwen3.8-27B, local coverage drops from 55.9 to 41.5\% between the two tiers, much more than compositional accuracy does (\cref{fig:read}b), and GLM-5.3-Flash and  DeepSeek-V4.1-Flash behave similarly.

\section{Analysis}
\label{sec:analysis}

The results share one shape: models perform well on what a print shows and degrade as a task demands more of what printing discarded. In particular, GPT-6~Astra scores 93 to 95\% in understanding and reaches 91.7 to 95.5 F1 on what is printed, yet only 62\% of the points it places correctly sit on the correct layer (\cref{fig:qual}). This demand is fair as a drafter assigns the nine layers from what an entity depicts, and the plot preserves line weight and type. Redrawing a line and deciding what it is for remain different acts, and in practice a file with the right lines on the wrong layers prints correctly but edits badly. The gap persists under more feedback, since a script that flags entities off the nine layers or with the wrong line type moves the L1 score of the GPT-6 models by only $-0.1$ and $+0.6$.

Across a set, what limits the models is how they spend effort. A set holds 5 to 14 pages, yet a set run takes only 1.2 to 2.5 times the steps of a single-page run, so each page receives a fraction of its solo effort. The loss then falls on what is written more than on what is drawn: from L1 to L2, GPT-6~Astra loses little geometry F1 but far more text and dimension F1 (\cref{sec:s-cases}), and the loss grows with set size, from a small drop at up to 7 pages to a large one at 11 or more. No model appears to plan its effort over a set.

Design demands the most and fails most clearly. The two GPT-6 models remove 82 and 81\% of the reference values that must go but write only 48 and 49\% of the new ones correctly (\cref{sec:s-l3}), so they know what a change touches but miscompute what it becomes. In the abutment-cap task of \cref{fig:design}, GPT-6~Astra recomputes two bar counts but keeps the plan rectangular, derives 7902 instead of 8712\,mm for bar 3, and leaves bar 4 unchanged, which makes every total wrong. Design values depend on one another through the drawing's rules, since counts follow from width and spacing, lengths from geometry, and totals from both, so one early error spreads downstream and every wrong edit loses a value that copying would have kept.

Finally, reading a drawing predicts little about producing one. DeepSeek-V4.1-Flash answers fewer Single-Sheet questions than GPT-5.6~Sol (60.4 against 72.2\%) yet rebuilds a page far better (81.3 against 59.8), plausibly because a question needs one fact while a file needs every entity, style, and value.

\section{Conclusion}
\label{sec:conclusion}

We introduced \ours, a benchmark of the drawing work behind civil infrastructure, built from real projects with the designers' source files as ground truth.
Its central lesson is that a print is a lossy projection of a file: the best models read the print and redraw a page accurately, but fall short of recovering layers, the consistency of a set and the dependence of values on parameters.
A reconstruction that prints like the original thus still needs checking, and since every ground truth is paired with its plot, the missing abilities can be trained without manual labels.

Future work will extend \ours to more structure types, disciplines and drafting standards, and from clean plots to scanned and marked-up prints.
The scorer compares a model's file with the designers' file and does not check structural adequacy or code compliance.

\bibliographystyle{unsrtnat}
\bibliography{main}

\beginappendix
The main text argues three things: that \ours measures drawing work against the designers' own files, that its scorer can be trusted, and that frontier models fail on what printing discards.
This appendix follows the same order.
\Cref{sec:s-bench} shows how the benchmark is built and checked, \cref{sec:s-scoring} describes the scorer and the faults that tested it, \cref{sec:s-results} opens up the columns of \cref{tab:main}, and \cref{sec:s-analysis} reports the controls that test the causes proposed in \cref{sec:analysis}.
\Cref{sec:s-tasks} closes with examples of what a model receives and produces, which the earlier sections refer to.
Each section begins by saying which part of the main text it supports.

\section{The benchmark in detail}
\label{sec:s-bench}

This section expands \cref{sec:gt}.
We first place \ours among related benchmarks and then describe how its pages and items were produced and checked; examples of the resulting tasks follow in \cref{sec:s-tasks}.

\subsection{Related benchmarks}
\label{sec:s-related}

\Cref{tab:compare} compares \ours with the closest benchmarks; here we describe what each of them asks of a model, so that the columns of the table read as differences between tasks.
Most existing work evaluates how well a model reads drawings.
MechVQA covers ten task types on mechanical drawings and trains a domain model on them~\cite{mechvqa}, and DrawingVQA writes questions with practitioners on construction drawings and reports a large gap to professionals~\cite{drawingvqa}.
AECV-Bench counts and reads elements on architectural plans~\cite{aecvbench}, MMArch and ArchSIBench test architectural evidence and spatial reasoning~\cite{mmarch,archsibench}, and MechReason asks multi-image, multi-hop questions on mechanical engineering material~\cite{mechreason}.
LED-BENCH, concurrent with this work, scores large-format sheets from three disciplines with an LLM jury~\cite{ledbench}, and AEC-Bench evaluates coding agents across intra-sheet, drawing and project scopes~\cite{aecbench}.
DocScope comes from the long-document literature and grades the evidence behind each answer~\cite{docscope}, a practice that our understanding tiers follow.

A second group asks models to write CAD.
Recent benchmarks evaluate programmatic CAD~\cite{benchcad,cadbench}, parametric authoring from engineering drawings~\cite{paramcadv3}, the engineering validity of generated designs~\cite{cadengbench} and long-horizon use of CAD software~\cite{cadworld}.
EngiWorld evaluates agents across 26 professional engineering applications and checks the artifacts they deliver with programmatic verifiers~\cite{engiworld}, and DrawAI makes general raster images editable with an agentic workflow and judges them with a VLM~\cite{drawai}.
These works share our protocol of scoring a produced artifact, but most of them target 3D parts or general images and not layered civil drawing sets.

Work in the civil domain is narrower.
TriView2CAD reads fifteen parameters from three-view drawings of prefabricated bridge piers~\cite{creftcad}, an LLM agent scripts AutoCAD to draw beam sections from a text description~\cite{structdrawagent}, and StructureClaw evaluates agents on structural analysis workflows through the artifacts they produce~\cite{structureclaw}.
Pipelines that reconstruct building information models from the 2D drawings of existing bridges and buildings~\cite{akanbi2022,zhao2021bim,zhang2021bimreview} document the need for reconstruction in practice, which is the situation that L1 and L2 mimic.

\subsection{From CAD files to pages}
\label{sec:s-data}

Every reconstruction and design page follows one rule: the input and the ground truth come from the same source file, and nothing in the input reveals what plotting dropped.
Each paper-space layout is flattened into paper millimetres on an A3 sheet ($420\times297$\,mm, origin at the lower left), blocks are bound, text encodings are normalised and the title block is removed, which gives the ground-truth DXF of one page.
The input is a monochrome plot of the same file in which each entity keeps the line weight and line type of the source drawing, rasterised at 200\,dpi into a PDF without vector or text layers.
These attributes are stored on the entity, so its layer does not affect the plot, and this is why layer assignment is the part of the file that printing discards (\cref{fig:samepix}).

Three further choices define the task levels.
Size tiers are set by entity count (\cref{fig:data}d), L2 sets are formed from each project's drawing numbers so that the pages of a set belong together, and L3 keeps the tasks whose changed values can be traced to the new parameters or to the pages given, with the reference DXF given after unreferenced blocks are removed.
Together they keep every task answerable in principle from what the model receives.

\subsection{Writing the understanding items}
\label{sec:s-items}

The understanding items implement the principle of \cref{sec:read} that a wrong choice should say which binding failed.
Every wrong option is what one named misbinding would produce and is refuted by the drawing; at most two options concern the same attribute, and the stem states the task without naming the trap.
For multi-sheet parents, the author records, per option, which single page could decide it, and decoy pages are same-format sheets of a different instance or competing tables.
Each parent comes with probes of two kinds: \textit{read} probes name the object and ask for a value, while \textit{bind} probes refer to the object only through a property printed on another sheet, and probe scorers are typed (a number with its unit, named slots, or endpoint pairs).

Before an item reaches an expert it passes two checks.
A compiler rejects items whose stem contains a value from a correct option and probes whose gold value appears in a parent option, and correct letters are shuffled after authoring so that each letter carries about one sixth of the correct options.
A second LLM agent that had not written the item then reviews it against the drawing, and only items that pass go on to the experts.

\subsection{Expert review}
\label{sec:s-experts}

Three civil engineering experts with graduate-level training reviewed every item against the drawing.
For each item they checked that the stem is unambiguous, that every correct option follows from the drawing alone, that every wrong option is refuted by it, and that no option depends on codes or assumptions outside the drawing.
For probes they checked that each one can be understood on its own and does not reveal the answer of its parent.
The experts cross-checked one another's reviews, and items on which they disagreed were discussed and then revised or removed, so that every item in the benchmark has been judged against the drawing by a person and not only by the agents that drafted it.

\section{Scoring in detail}
\label{sec:s-scoring}

This section supplements \cref{sec:scoring}.
We first state the matching rule in full and then describe the faults that we used to test the scorer.

\subsection{Matching and missing outputs}
\label{sec:s-matching}

Text F1 and dimension F1 both use the matching rule of \cref{sec:scoring}.
A text may match a dimension label and vice versa, because some source pages store dimensions as exploded lines and text, and dimension F1 therefore scores the labels, their values and positions, and not the entity type.
Shared-value recall and design recall compare a page's text and dimension values as strings, wherever they are placed, because they test the values and the layout is already covered by geometry F1.
A missing or unreadable output scores 0, and in L2 missing pages score 0.
Where the layer mapping does not cover a page, its score is the weighted mean of the remaining components.

\subsection{Injected faults}
\label{sec:s-faults}

To show that the scorer penalises what matters and ignores what does not (\cref{fig:samepix}e), we injected faults into the ground truths and scored them against the originals.
Two faults change only layers: moving all entities to one layer, and swapping the outline and reinforcement layers.
Three delete 10\% of the lines, of the texts or of the dimensions, and two change one digit in 10\% of the texts or of the dimension values, so that content is lost or corrupted.
One jitters 10\% of the lines by 3\,mm, and one shifts the whole page by 3\,mm, which changes no content and tests the registration.
Every content fault lowered the score, the page shift was largely absorbed (98.6), and the two layer-only faults cost the most (81.0 and 86.7) although they leave every printed pixel unchanged.

\section{Full results}
\label{sec:s-results}

\Cref{tab:main} condenses seven models, three tracks and several scales of difficulty into a few columns.
This section opens up the three columns that need more detail: understanding by binding type and by probe, the structure of the design errors, and the effort that the runs take.

\subsection{Understanding by binding type and probe}
\label{sec:s-read}

\Cref{tab:s-read} splits the Single-Sheet accuracy by binding type and adds the probe accuracy of the two multi-sheet tiers.
Three patterns stand out.
First, table lookups are the easiest type for every one of the seven models, while types that tie a number to a drawing convention are harder: for the GPT-6 models, view and scope items are the lowest at 85.7 to 87.8\%, and datum items drop to 48.1\% for Qwen3.8-27B and 23.8\% for Qwen3.5-9B.
Second, resolution matters for models with a small image budget: tiling each page lifts DeepSeek-V4.1-Flash from 50.8 to 60.4\% on Single-Sheet items and from 43.1 to 58.7\% on Full-Set parents, which is why the main text reports the tiled input.
Third, probes are answered more accurately than the parents they decide, except for GPT-6 Astra on Few-Sheet, so a model that reads each fact often still fails to assemble all the facts of a parent; Qwen3.5-9B marks the other extreme, with only 8.2\% of Full-Set parents and 16.0\% of their probes answered correctly, which shows that it already fails to find and read the single facts.

\begin{table*}[t]
\centering
\caption{\textbf{Understanding accuracy by binding type and for probes.} Single-Sheet items and parents are multi-select questions with six options scored by exact match (chance 1.8\%); probes are short-answer questions. Single-Sheet columns split the 1{,}800 items by binding type. $\dagger$: each page is followed by its $2{\times}2$ tiles, the input reported in the main text.}
\label{tab:s-read}
\small
\setlength{\tabcolsep}{2.9pt}
\begin{tabular}{l c cccccc cc cc}
\toprule
 & \multicolumn{7}{c}{Single-Sheet} & \multicolumn{2}{c}{Few-Sheet} & \multicolumn{2}{c}{Full-Set} \\
\cmidrule(lr){2-8}\cmidrule(lr){9-10}\cmidrule(lr){11-12}
Model & All & Table & Instance & Cross-view & Datum & View & Scope & Parent & Probe & Parent & Probe \\
\midrule
GPT-6~Astra~\cite{gpt6astra} & \textbf{93.3} & 95.7 & \textbf{94.2} & \textbf{92.5} & \textbf{94.5} & \textbf{87.8} & \textbf{87.0} & \textbf{95.3} & \textbf{93.8} & \textbf{94.5} & \textbf{95.6} \\
GPT-6.1~Sol~\cite{gpt61sol} & 91.5 & \textbf{98.4} & 91.9 & 89.7 & 91.3 & 85.7 & 85.9 & 92.2 & 93.2 & 89.7 & 94.1 \\
GPT-5.6~Sol~\cite{gpt56sol} & 72.2 & 87.2 & 71.7 & 67.4 & 71.9 & 76.9 & 62.8 & 75.5 & 83.0 & 69.4 & 85.4 \\
\midrule
DeepSeek-V4.1-Flash$^\dagger$~\cite{deepseekv41flash} & 60.4 & 63.5 & 63.1 & 57.6 & 58.4 & 50.6 & 53.4 & 63.8 & 81.7 & 58.7 & 80.6 \\
DeepSeek-V4.1-Flash~\cite{deepseekv41flash} & 50.8 & 58.9 & 51.3 & 48.9 & 50.7 & 44.6 & 43.8 & 52.1 & 60.8 & 43.1 & 62.3 \\
Qwen3.8-27B~\cite{qwen38} & 58.2 & 75.4 & 60.2 & 51.9 & 48.1 & 57.1 & 52.1 & 61.4 & 82.4 & 53.6 & 75.7 \\
GLM-5.3-Flash~\cite{glm53flash} & 57.4 & 61.9 & 58.7 & 55.4 & 56.2 & 48.0 & 55.4 & 60.2 & 78.4 & 52.8 & 77.9 \\
Qwen3.5-9B~\cite{qwen35} & 34.3 & 55.6 & 34.2 & 31.5 & 23.8 & 24.2 & 34.0 & 43.1 & 67.1 & 8.2  & 16.0 \\
\bottomrule
\end{tabular}
\end{table*}

\subsection{The structure of the design errors}
\label{sec:s-l3}

The L3 results of \cref{tab:main} mix two things: how many values a model changes, and how many it changes correctly.
The stale rate, the share of values that should change but stay at their reference value, separates them; it is 0.18 for GPT-6 Astra, 0.19 for GPT-6.1 Sol and 0.34 for GPT-5.6 Sol.
Together with the design recall of 48, 49 and 40\%, it splits the values that should change into three shares: GPT-6 Astra leaves 18\% at the reference, writes 48\% correctly and gets or leaves out the remaining 34\%, GPT-6.1 Sol does the same with 19, 49 and 32\%, and GPT-5.6 Sol with 34, 40 and 26\%.
GPT-5.6 Sol is thus the more conservative model, whereas the GPT-6 models change more values and are wrong about a third of them.
The page scores follow from this: the mean is 58.1 for GPT-6 Astra, 55.2 for GPT-6.1 Sol and 65.1 for GPT-5.6 Sol, against 64.1 for copying the reference page.
Task by task, GPT-5.6 Sol stays at the level of the copy because it tends to leave the reference page unchanged, which is why the main text reports design recall next to the page score.

\subsection{Effort per run}
\label{sec:s-effort}

Effort is a possible confound, since a model might score higher simply because it works longer.
\Cref{tab:s-effort} reports the steps, output tokens and input tokens of a run, and two observations argue against this explanation.
GPT-5.6 Sol spends about as much on an L1 page as GPT-6 Astra (21 against 22 steps, 20.6k against 22.1k output tokens) yet scores 31 points lower, and the three open models in the table take an order of magnitude more steps per page (224 to 432 against 21 to 31 for the GPT models) and still trail.
A set run takes only 1.2 to 2.5 times the steps of a page run (42 against 22 for GPT-6 Astra), which is the observation behind the effort claim of \cref{sec:analysis}.
Qwen3.8-27B also takes the fewest steps on L3 relative to L1 (51 against 432), and its design recall is only 6\%.

\begin{table*}[t]
\centering
\caption{\textbf{Effort per reconstruction or design run.} Mean per run with the median in parentheses; an L2 run rebuilds one set. Steps are the agent's tool calls. Output tokens (Out, thousands) include reasoning tokens. Input tokens (In, millions) count the full context sent with every request, most of it read from cache.}
\label{tab:s-effort}
\footnotesize
\setlength{\tabcolsep}{2.1pt}
\begin{tabular}{@{}l ccc ccc ccc@{}}
\toprule
 & \multicolumn{3}{c}{L1: rebuild a page} & \multicolumn{3}{c}{L2: rebuild a set} & \multicolumn{3}{c@{}}{L3: design a page} \\
\cmidrule(lr){2-4}\cmidrule(lr){5-7}\cmidrule(l){8-10}
Model & Steps & Out (k) & In (M) & Steps & Out (k) & In (M) & Steps & Out (k) & In (M) \\
\midrule
GPT-6~Astra~\cite{gpt6astra} & 22 (20.5) & 22.1 (21.6) & 1.27 & 42 (41.5) & 37.4 (35.9) & 2.48 & 18 (19) & 16.8 (15.3) & 1.37 \\
GPT-6.1~Sol~\cite{gpt61sol} & 31 (29) & 30.3 (29.8) & 2.14 & 52 (53.5) & 51.8 (48.5) & 3.72 & 24 (24) & 22.6 (21.3) & 1.91 \\
GPT-5.6~Sol~\cite{gpt56sol} & 21 (19) & 20.6 (20.1) & 1.00 & 25 (24.5) & 24.7 (24.7) & 1.43 & 27 (28) & 23.8 (22.8) & 2.02 \\
\midrule
DeepSeek-V4.1-Flash~\cite{deepseekv41flash} & 256 (259) & 238.3 (231.4) & 21.5 & 559 (558.5) & 546.6 (518.9) & 50.9 & 154 (136) & 207.4 (191.6) & 12.8 \\
Qwen3.8-27B~\cite{qwen38} & 432 (416) & 398.6 (389.5) & 35.4 & 734 (712) & 842.8 (832.1) & 83.2 & 51 (51) & 136.5 (150.6) & 7.44 \\
GLM-5.3-Flash~\cite{glm53flash} & 224 (219.5) & 263.5 (268.8) & 18.3 & 561 (577.5) & 922.8 (975.8) & 47.9 & 123 (121) & 222.9 (240.0) & 10.2 \\
\bottomrule
\end{tabular}
\end{table*}

\section{Controls and further analysis}
\label{sec:s-analysis}

The analysis of \cref{sec:analysis} makes claims about causes.
This section tests two alternatives that a reader may raise: that the agent loop is the bottleneck, and that a model cannot read what it has produced.

\subsection{Does the agent loop explain the layer gap?}
\label{sec:s-controls}

Two controls test whether the agent's own loop drives the L1 page score: removing the preview tool, and adding a script that checks the agent's file.
For the three GPT models, neither changes the score significantly on the 8 control pages: removing the preview tool changes it by $-1.8$ points (95\% interval $[-6.6, 3.3]$) and adding the script by $+2.6$ ($[-0.4, 6.8]$).
The control pages were fixed in advance by rule, with at most one page per drawing number.
In the open-loop condition the preview tool was removed, and the task statement forbade rendering the output or comparing it with the input.

The verifier script holds no reference data.
It lists file errors, entities on non-target layers or with the wrong line type, linear dimensions whose displayed value disagrees with their drawn length at the page's common scale, entities outside the sheet, and duplicated segments.
Its effect is $-0.1$ for GPT-6 Astra, $+0.6$ for GPT-6.1 Sol and $+7.5$ for GPT-5.6 Sol.
Only the weakest model benefits, which suggests that its errors are of the kind a checker can catch, whereas the errors of the two stronger models are not, and that the layer gap of \cref{sec:analysis} is not a matter of missing feedback.

\subsection{Reading from a reconstruction}
\label{sec:s-pairing}

The main text tests reconstruction and understanding separately, so a model could pass one and fail the other for unrelated reasons.
We ask the link directly: can a model answer a question from the file it produced instead of from the printed page?
The 122 items are all the understanding items that lie on one of the 92 reconstruction pages that carry such items.
The three GPT models answered each item in four conditions inside the same container: from the input raster (A), from their own DXF only (B), from the source DXF only (C), and from the question and options alone (F); \cref{tab:pairing} gives the accuracy in all four.

For the GPT-6 models the model's own DXF is as informative as the print: accuracy differs by 2.5 and 0.9 points between A and B, and about five items per model disagree.
For GPT-5.6 Sol it is not, since accuracy falls from 87.7 to 78.7\%, with 18 items right only from the pixels and 7 only from its DXF (exact McNemar test $p{=}.043$, Holm-corrected over the three models $.13$), a difference that is suggestive but not significant after correction.
Condition C shows that the information is in the source file, because accuracy from the source DXF equals or exceeds that from the print for the GPT-6 models.
Condition F, without any drawing, stays at 11.5 to 12.3\%, far above chance (1.8\%) but far below A, so the items do need the drawing.
Taken together, a strong reconstruction keeps what the questions need, and the weaker one loses part of it.

\begin{table}[t]
\centering
\caption{\textbf{Paired accuracy (\%) on 122 understanding items.} A: input raster. B: own DXF. C: source DXF. F: no drawing.}
\label{tab:pairing}
\small
\begin{tabular}{lcccc}
\toprule
Model & A & B & C & F \\
\midrule
GPT-6~Astra~\cite{gpt6astra} & 95.9 & 93.4 & 96.7 & 11.5 \\
GPT-6.1~Sol~\cite{gpt61sol} & 93.4 & 94.3 & 95.9 & 12.3 \\
GPT-5.6~Sol~\cite{gpt56sol} & 87.7 & 78.7 & 86.9 & 12.3 \\
\bottomrule
\end{tabular}
\end{table}

\section{Examples: what a model receives and produces}
\label{sec:s-tasks}

The main text describes the tasks in words, and the previous sections measure and analyse them; this last section shows them.
We start with the understanding items, then turn to the reconstruction and design tasks, follow one reconstruction run from input to score, and end with three case studies that connect the aggregate results of \cref{sec:s-results} to single outputs.

\subsection{Understanding items}
\label{sec:s-examples}

\Cref{fig:examples} shows one Single-Sheet item for each binding type as the model receives it: a full page and a multi-select question with six options.
The six types are the binding points of \cref{sec:read}, and each wrong option corresponds to one misbinding of its type.
\Cref{fig:examples-multi} shows a multi-sheet parent under both access conditions, together with its probes.
Its correct option is the sum of three values that the read probes ask for, one per evidence page, and the decoy pages carry quantity tables of the same form whose totals the question excludes, so the probes show whether a wrong answer comes from misreading a value or from picking up a decoy.

\newcommand{\exitem}[4]{%
\begin{minipage}[t]{0.485\linewidth}
\centering
\setlength{\fboxsep}{0pt}\setlength{\fboxrule}{0.3pt}\fbox{#1}\par\vspace{0.45em}
\begin{minipage}{0.96\linewidth}\footnotesize
\textbf{#2.} #3\par\vspace{0.35em}
\centering #4
\end{minipage}
\end{minipage}}
\newcommand{\exopts}[6]{\begin{tabular}{@{}l@{\hspace{0.5em}}l@{\hspace{1.8em}}l@{\hspace{0.5em}}l@{\hspace{1.8em}}l@{\hspace{0.5em}}l@{}}
\textbf{A} & #1 & \textbf{B} & #2 & \textbf{C} & #3 \\
\textbf{D} & #4 & \textbf{E} & #5 & \textbf{F} & #6
\end{tabular}}
\begin{figure*}[p]
\centering
\exitem{\includegraphics[width=0.76\linewidth]{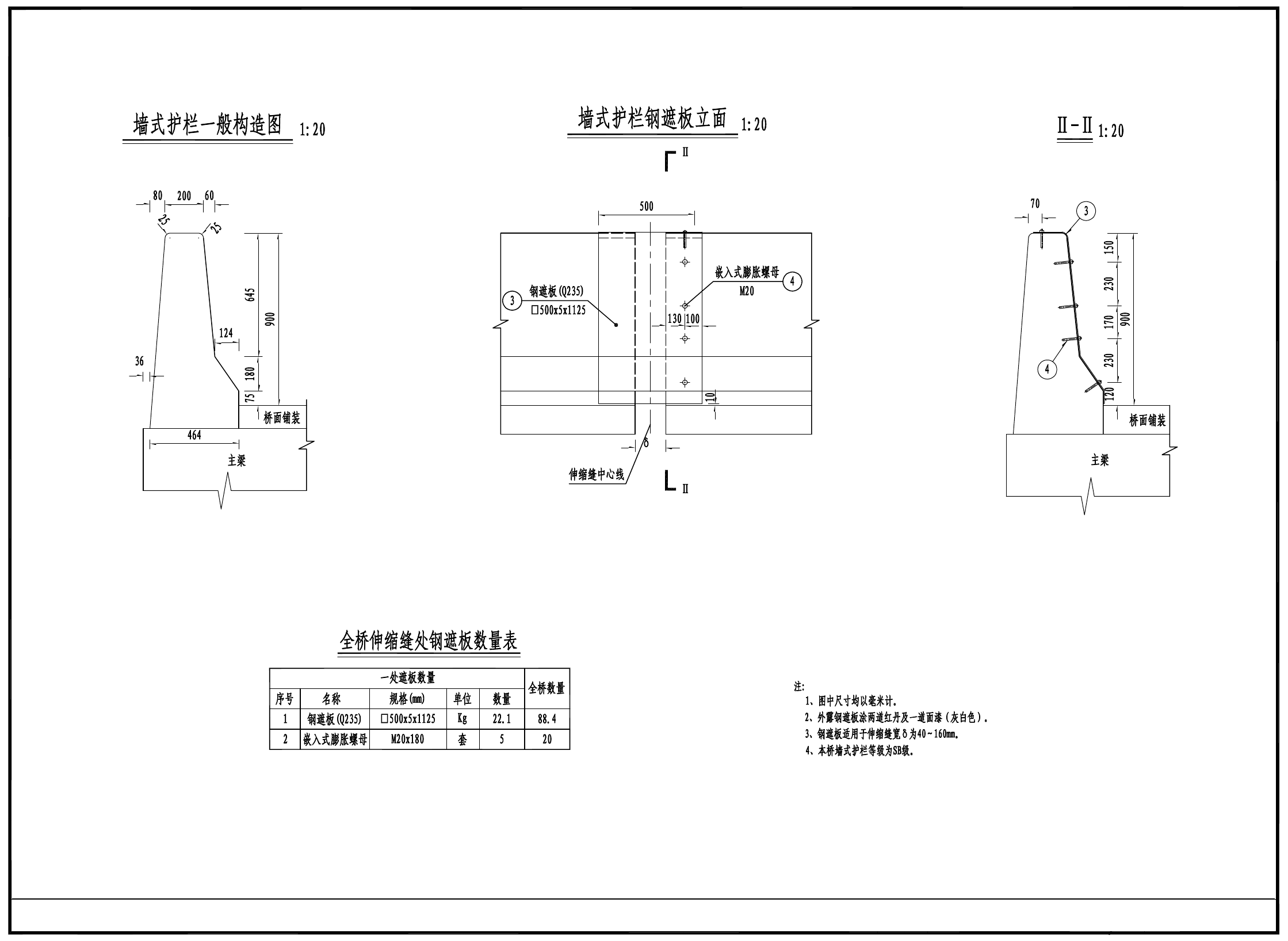}}{Datum}
{Which of the following are vertical distances from the top of the deck pavement to the centre of a side anchor of the barrier's steel cover plate (the anchor on top of the barrier excluded)? Units: mm.}
{\exopts{230}{120~\cmark}{900}{520~\cmark}{350~\cmark}{750~\cmark}}\hfill
\exitem{\includegraphics[width=0.76\linewidth]{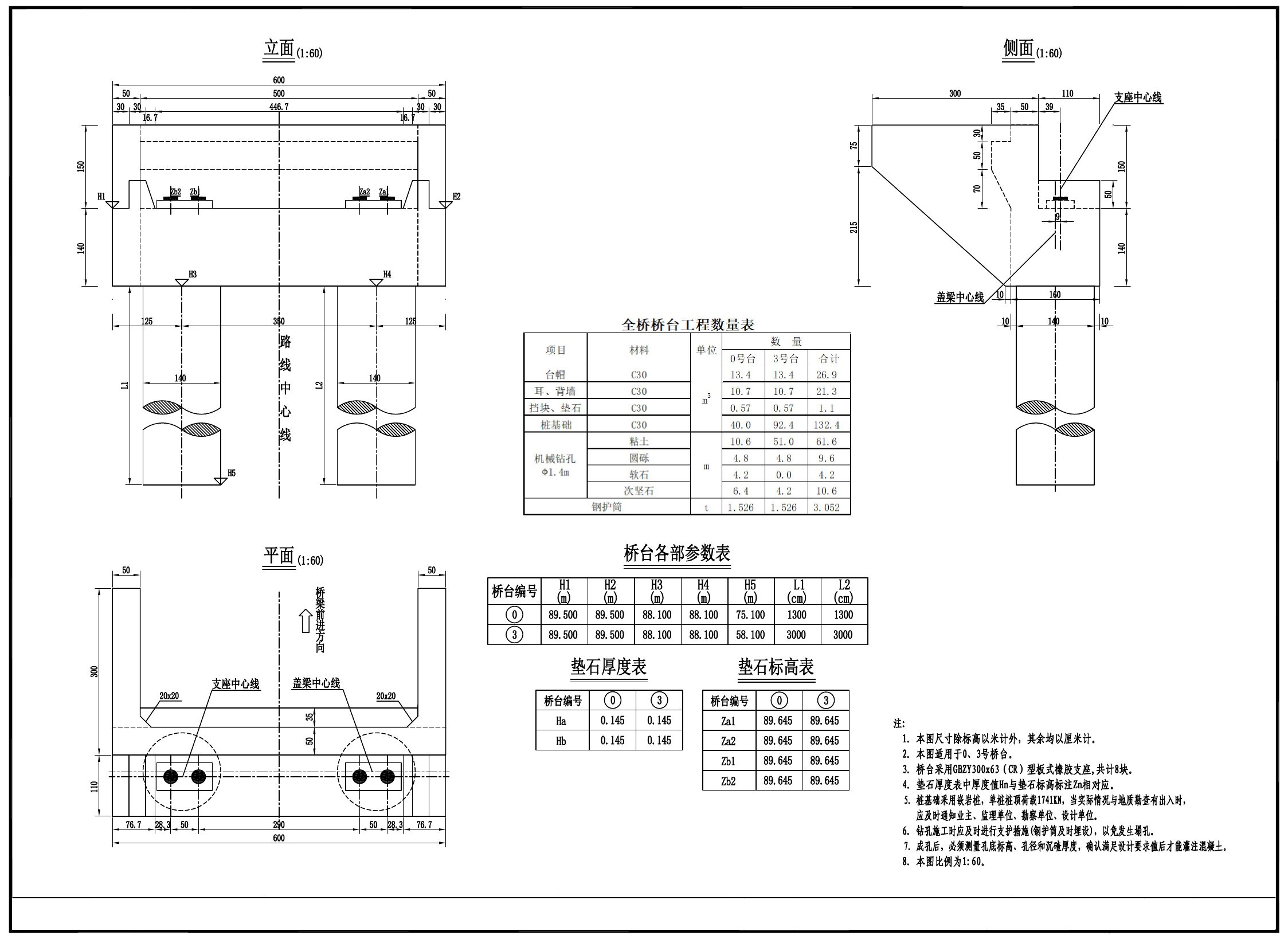}}{View}
{In the side view of the abutment, find the horizontal distances from the bearing centreline to the front edge of the abutment cap, from the bearing centreline to its rear edge, and between the bearing centreline and the cap-beam centreline. Which values are among the results?}
{\exopts{71\,cm~\cmark}{89\,cm~\cmark}{50\,cm}{110\,cm}{39\,cm}{80\,cm}}

\vspace{0.9em}
\exitem{\includegraphics[width=0.76\linewidth]{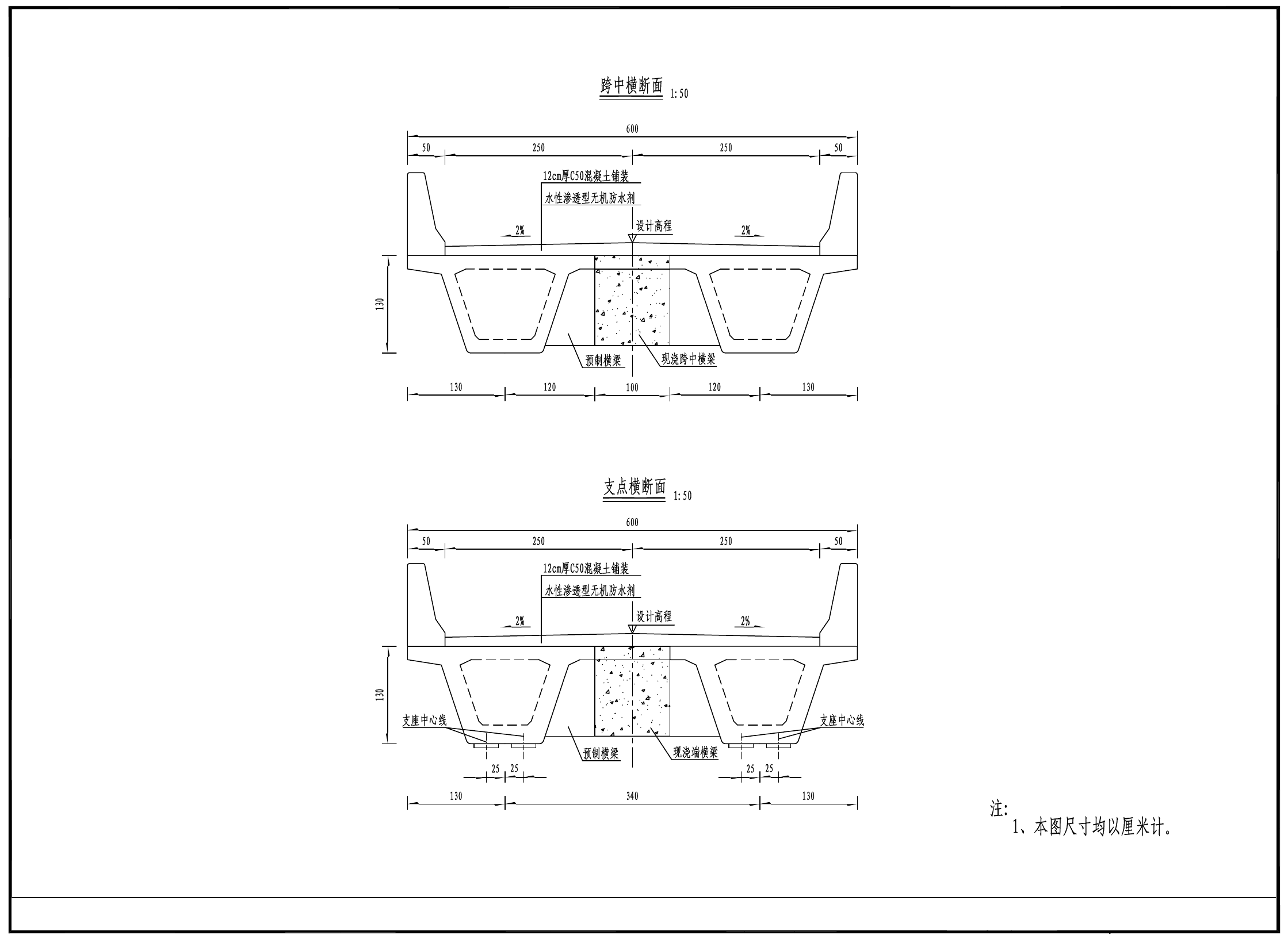}}{Cross-view}
{Taking the outermost left edge of the deck as the transverse origin, positive to the right, which values are transverse positions of the bearing centrelines in the cross-section at the support? Units: cm.}
{\exopts{130}{155~\cmark}{105~\cmark}{495~\cmark}{445~\cmark}{470}}\hfill
\exitem{\includegraphics[width=0.76\linewidth]{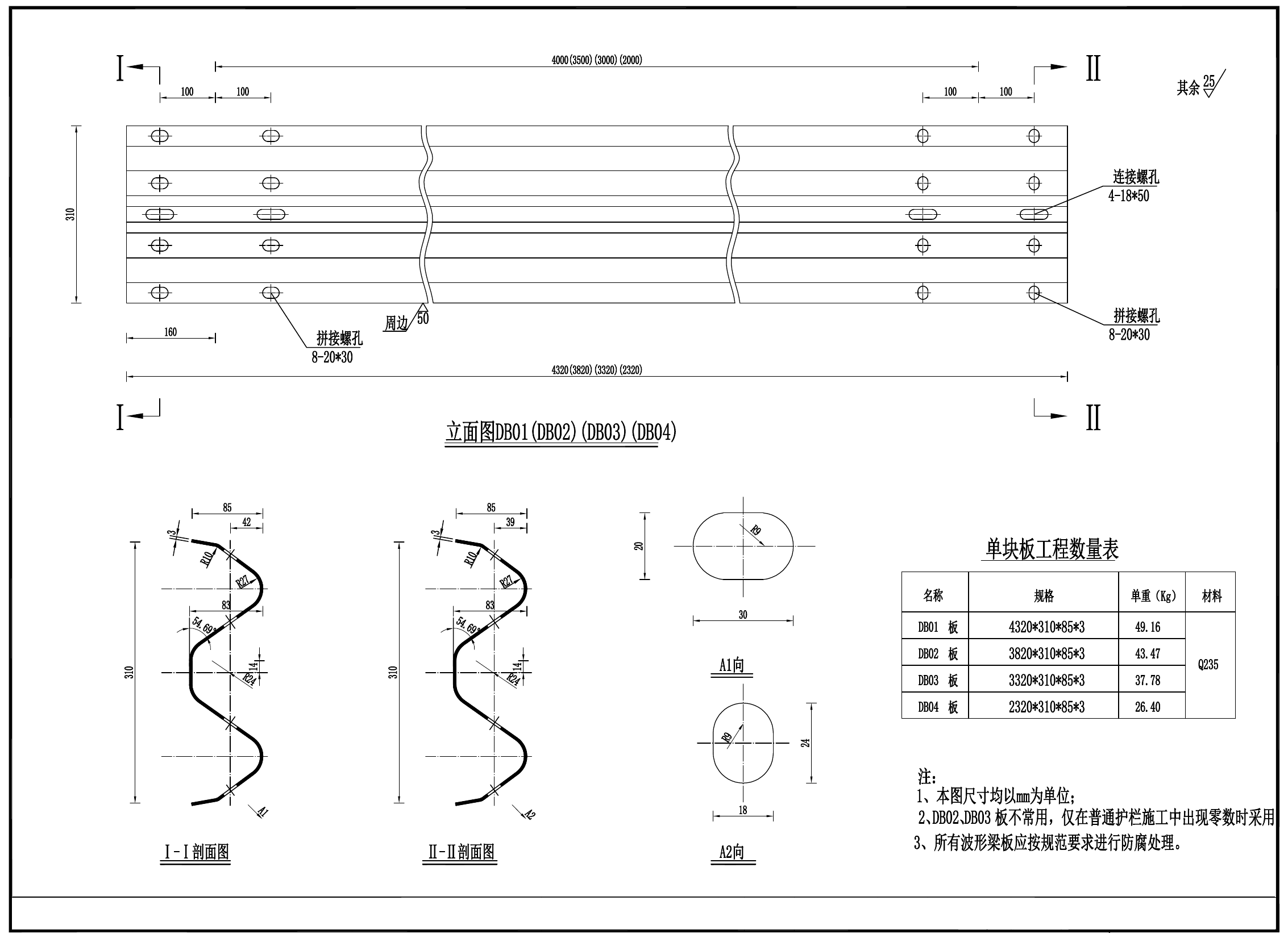}}{Instance}
{In sections I--I and II--II, take the horizontal distance from the cross mark on the uppermost inclined straight segment to the outermost edge of the wave crest to its right. Which values does the set of these distances contain? Units: mm.}
{\exopts{85}{42~\cmark}{39~\cmark}{83}{3}{14}}

\vspace{0.9em}
\exitem{\includegraphics[width=0.76\linewidth]{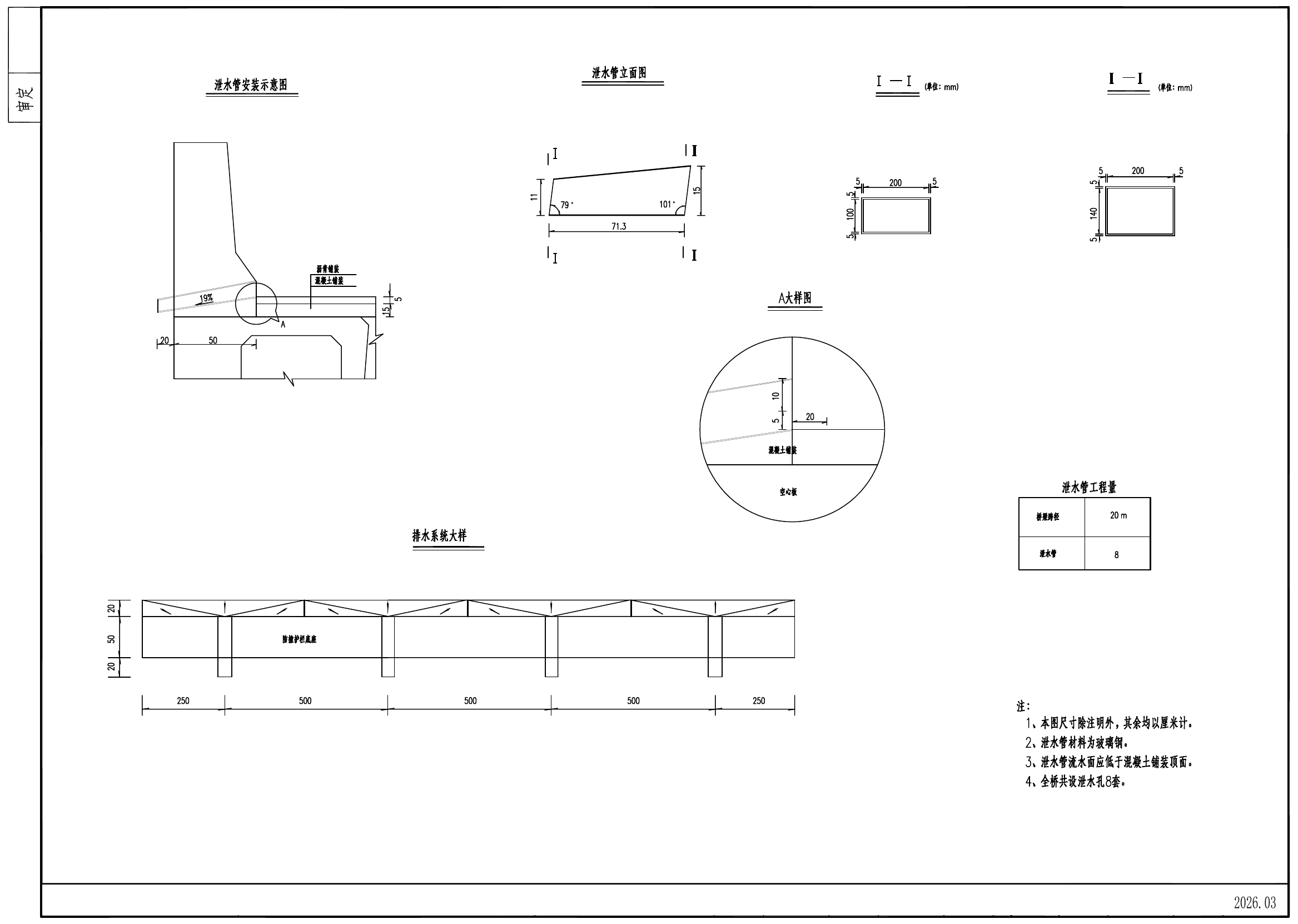}}{Scope}
{For the end faces I--I at the two ends of the drain pipe, which values are among the cross-sectional areas of the inner opening and of the solid pipe wall? Units: mm$^2$.}
{\exopts{31500}{20000~\cmark}{3500~\cmark}{28000~\cmark}{3100~\cmark}{23100}}\hfill
\exitem{\includegraphics[width=0.76\linewidth]{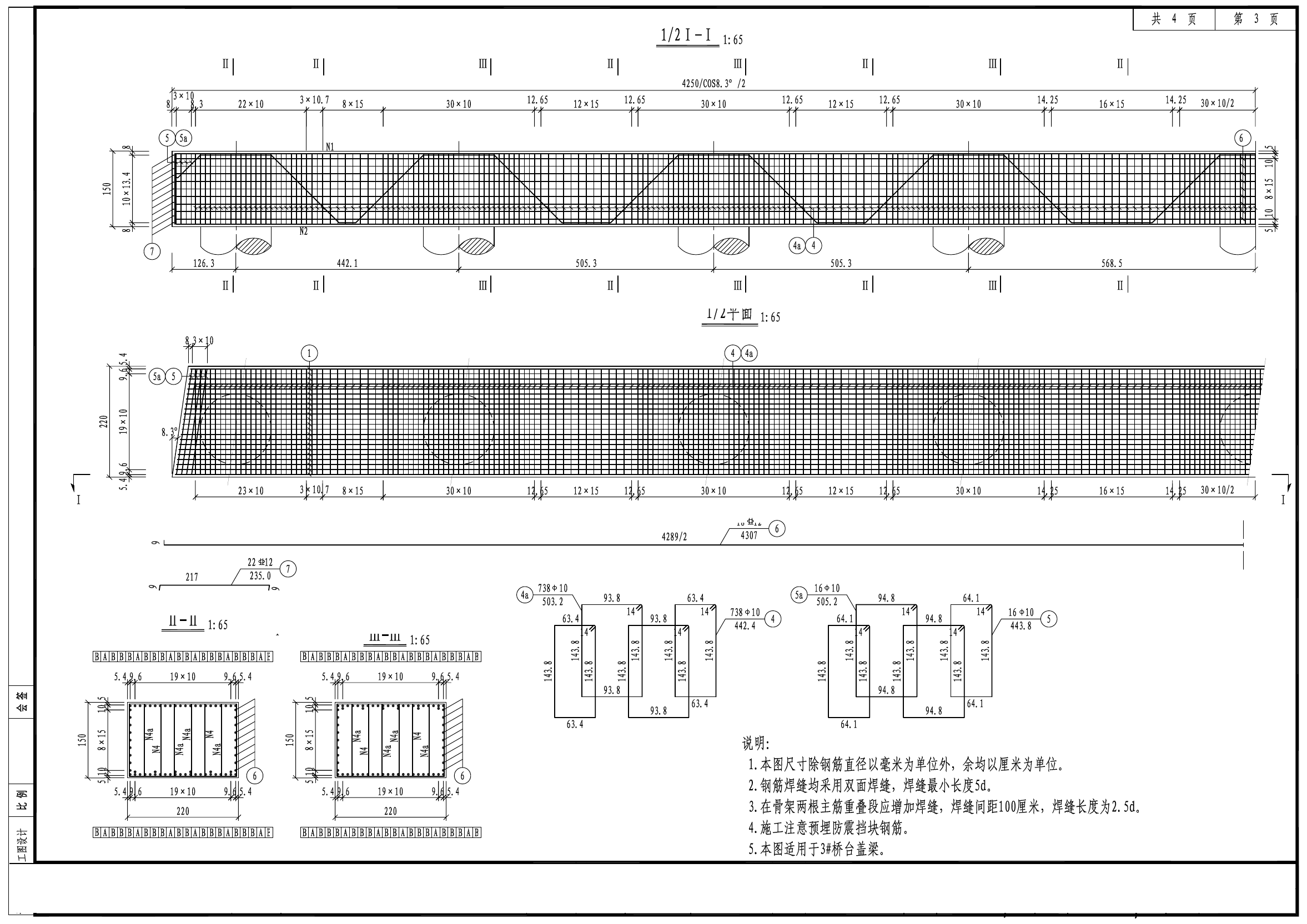}}{Table}
{Which numbered bars have a detail that is a stirrup with several right-angle bends and whose lowest horizontal segment is dimensioned longer than 90\,cm?}
{\exopts{N6}{N4}{N7}{N5a~\cmark}{N4a~\cmark}{N5}}
\caption{\textbf{Examples of Single-Sheet items, one per binding type.} The model receives the full page and a multi-select question with six options; \cmark{} marks the correct options.}
\label{fig:examples}
\end{figure*}

\newcommand{\exmpage}[2]{%
\begin{minipage}[t]{0.322\linewidth}\centering
\setlength{\fboxsep}{0pt}\setlength{\fboxrule}{0.3pt}\fbox{#1}\par\vspace{0.25em}
{\footnotesize #2}
\end{minipage}}
\begin{figure*}[p]
\centering\footnotesize
\begin{minipage}{0.94\linewidth}
\textbf{Parent question.} What is the total steel weight, welds included, of the type A, B and C segments of the stiffening girder over the whole bridge, excluding the longitudinal connections between segments and the wind-bearing stiffeners? Units: kg.\par\vspace{0.4em}
\centering
\begin{tabular}{@{}l@{\hspace{0.5em}}l@{\hspace{2.2em}}l@{\hspace{0.5em}}l@{\hspace{2.2em}}l@{\hspace{0.5em}}l@{}}
\textbf{A} & 13\,511\,864.31 & \textbf{B} & 13\,865\,116.48 & \textbf{C} & 14\,107\,950.13~\cmark \\
\textbf{D} & 14\,034\,551.85 & \textbf{E} & 13\,899\,458.41 & \textbf{F} & 13\,809\,907.22
\end{tabular}
\end{minipage}

\vspace{1.1em}
\begin{minipage}{\linewidth}\raggedright\textbf{(a) Few-Sheet:} the three evidence pages and two decoys, given in one turn in this order.\end{minipage}\par\vspace{0.45em}
\exmpage{\includegraphics[width=\dimexpr\linewidth-0.6pt\relax]{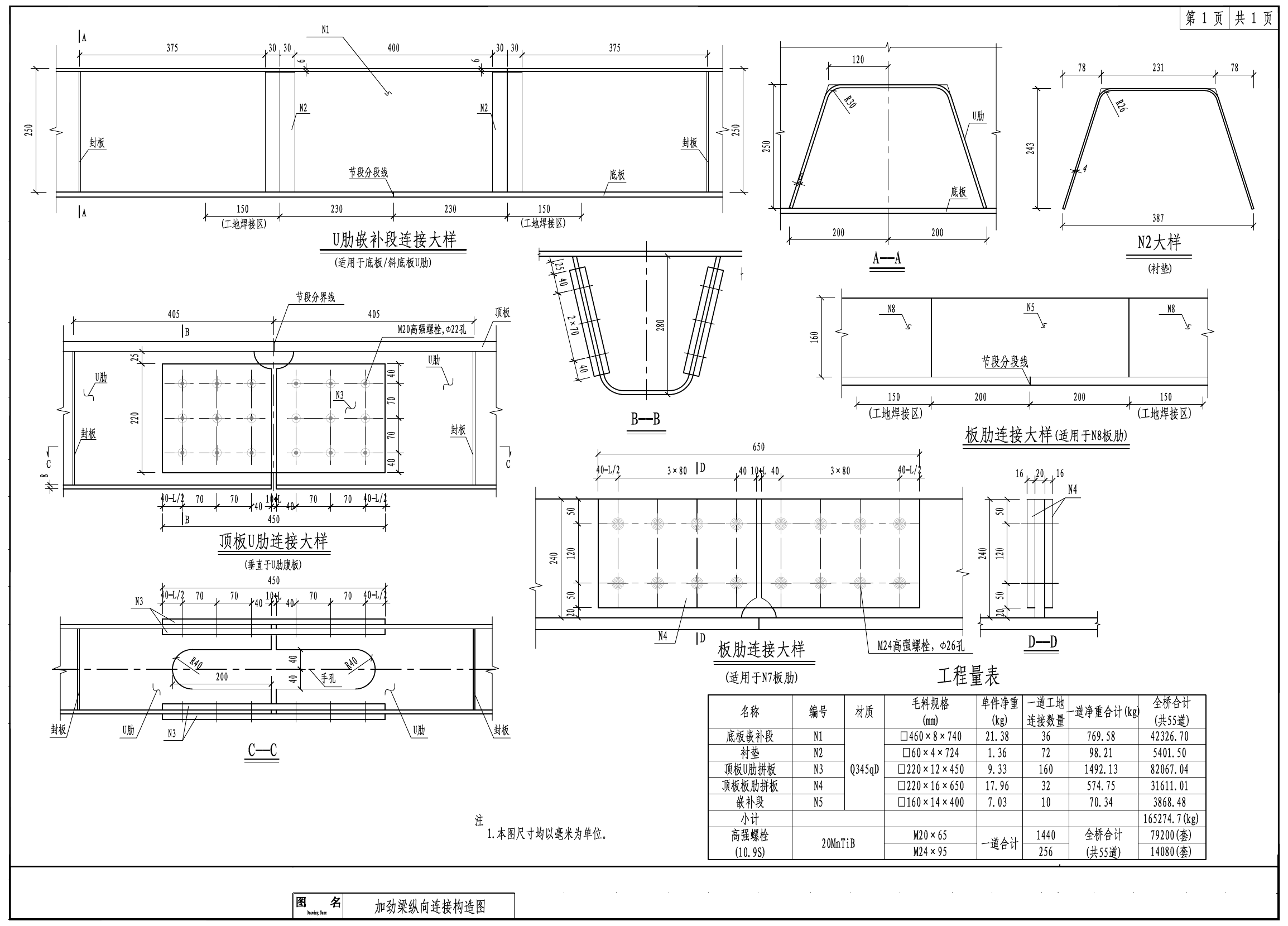}}{decoy: longitudinal connections}\hfill
\exmpage{\includegraphics[width=\dimexpr\linewidth-0.6pt\relax]{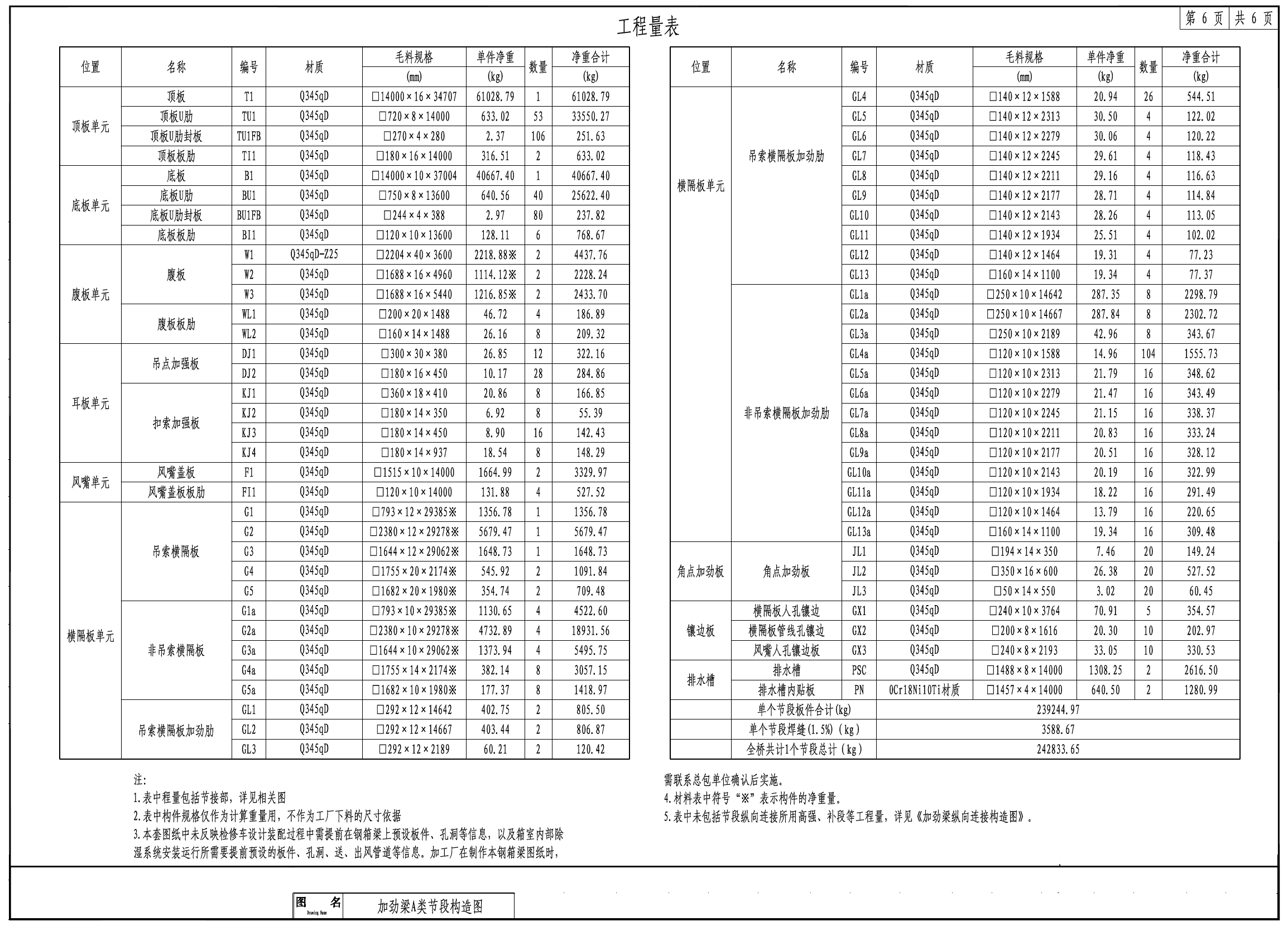}}{evidence: type A segment}\hfill
\exmpage{\includegraphics[width=\dimexpr\linewidth-0.6pt\relax]{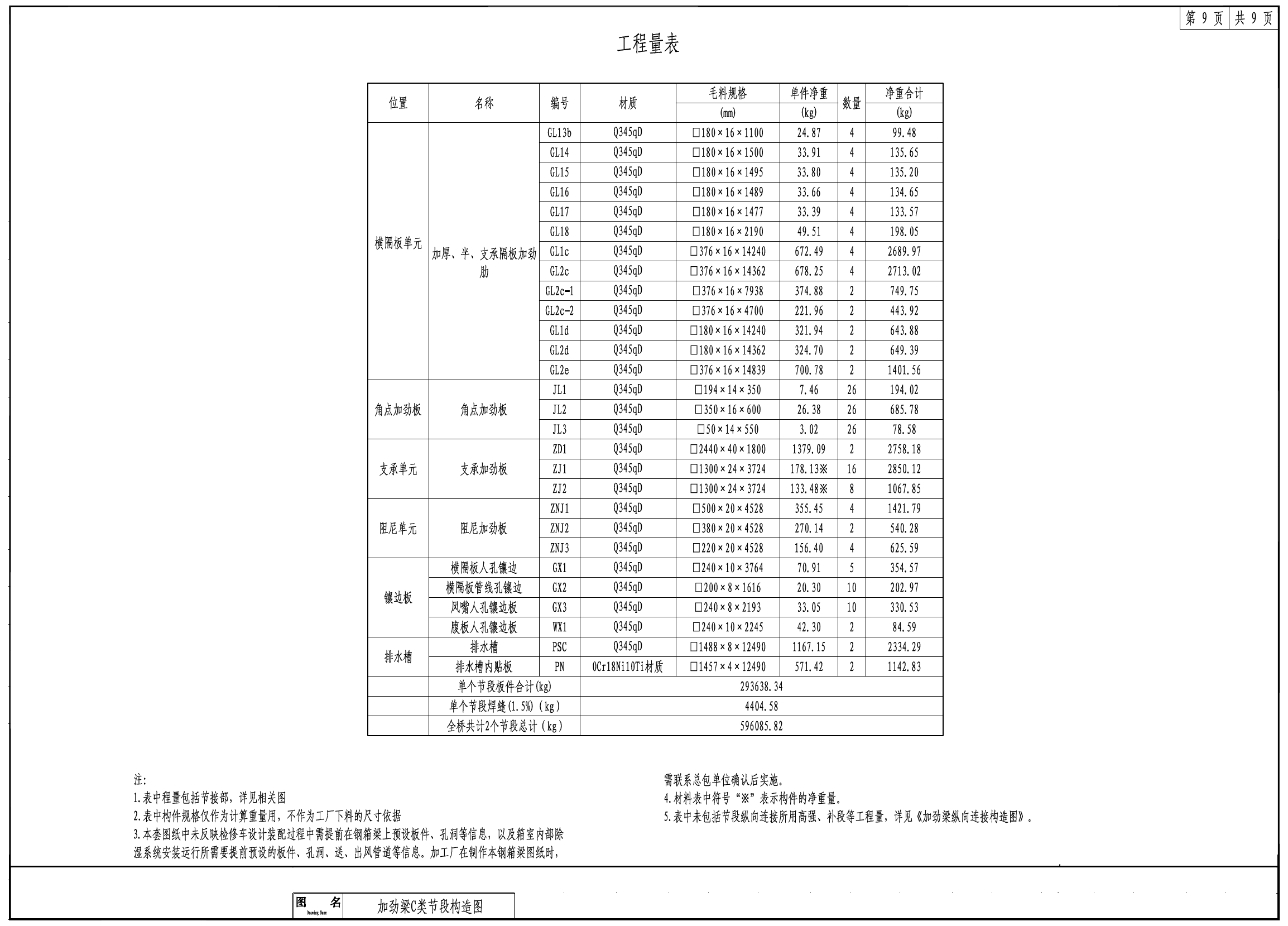}}{evidence: type C segment}

\vspace{0.7em}
\exmpage{\includegraphics[width=\dimexpr\linewidth-0.6pt\relax]{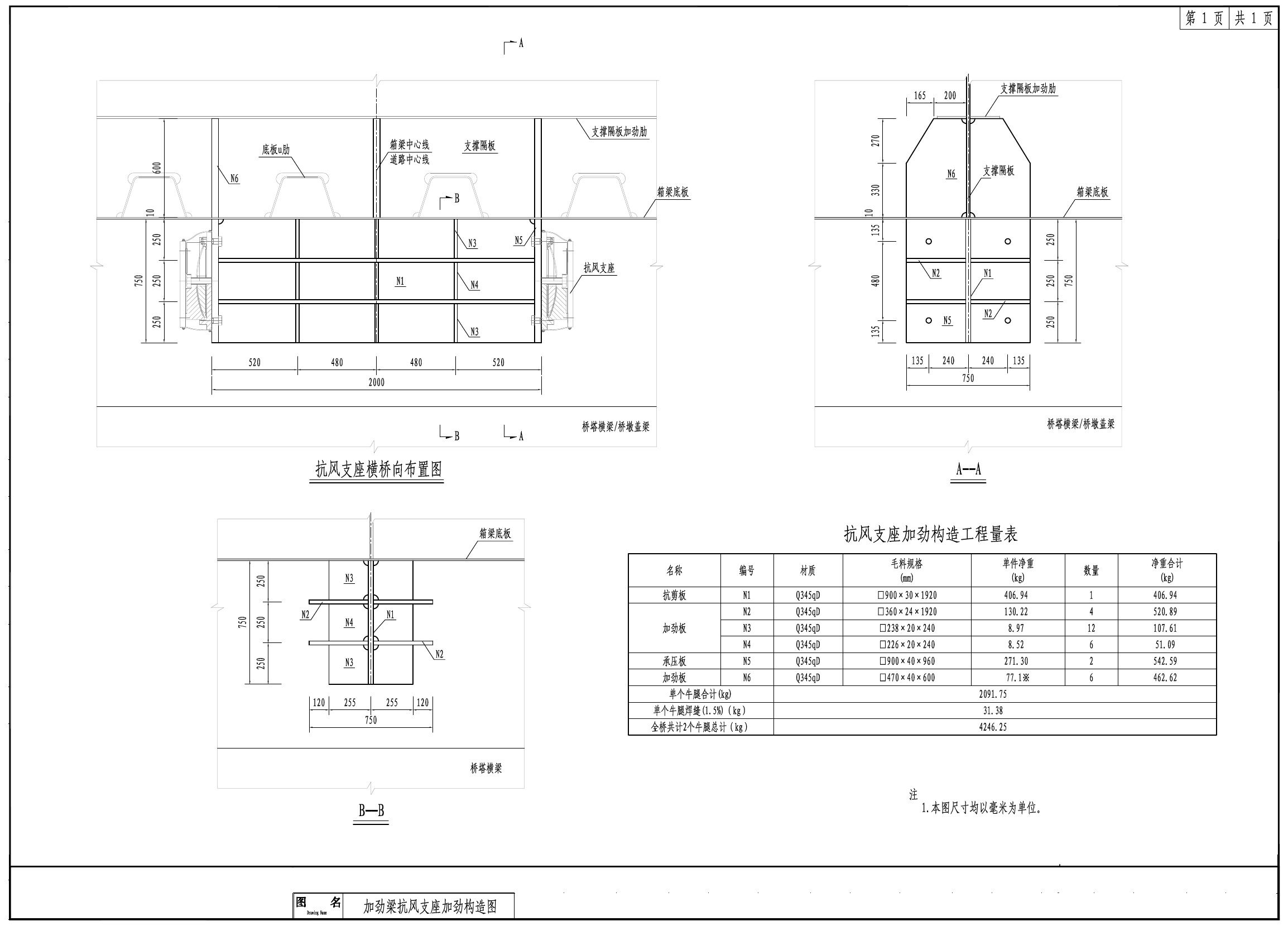}}{decoy: wind-bearing stiffeners}\hspace{0.017\linewidth}%
\exmpage{\includegraphics[width=\dimexpr\linewidth-0.6pt\relax]{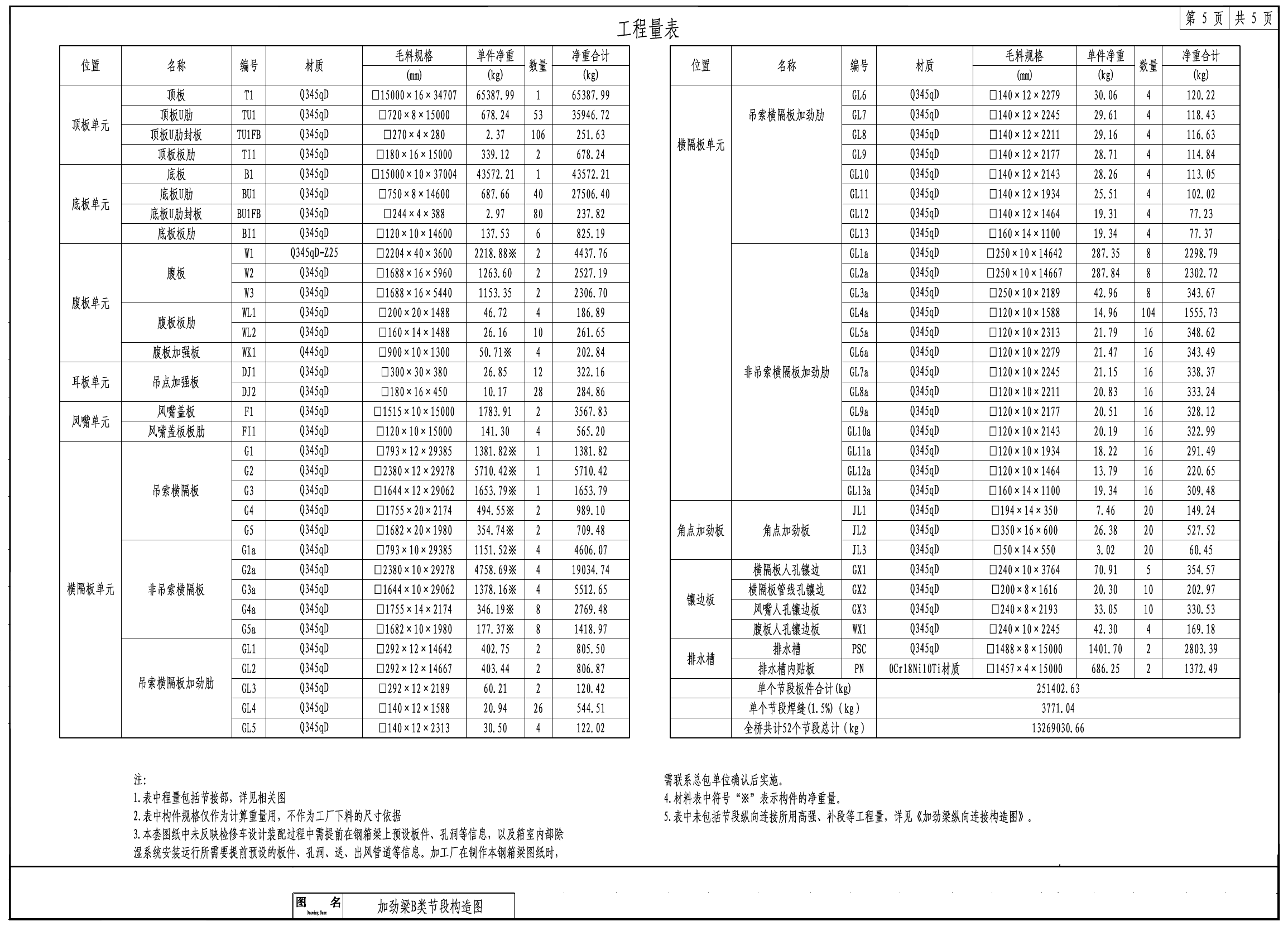}}{evidence: type B segment}

\vspace{1.1em}
\begin{minipage}{\linewidth}\raggedright\textbf{(b) Full-Set:} the same question over the whole unit of 27 pages, which the agent opens with \texttt{view\_page}.\end{minipage}\par\vspace{0.45em}
\includegraphics[width=\linewidth]{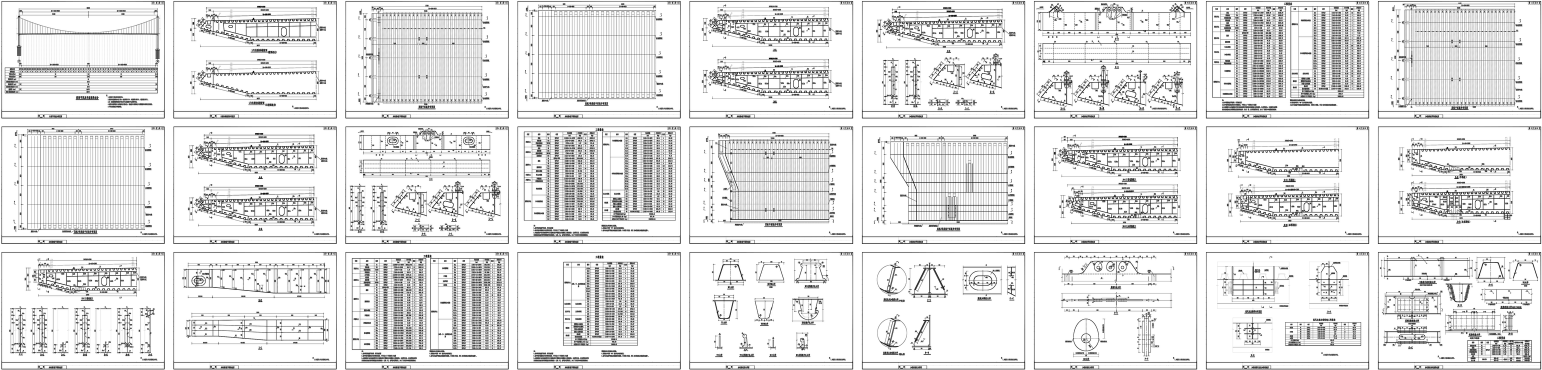}

\vspace{1.1em}
\begin{minipage}{\linewidth}\raggedright\textbf{(c) Probes} of this parent, asked under the same access condition.\end{minipage}\par\vspace{0.45em}
\resizebox{\linewidth}{!}{\begin{tabular}{@{}llr@{}}
\toprule
Role & Probe & Gold \\
\midrule
Read & Type A segment sheet, quantity table: what is the last row, the total over all segments of the bridge? & 242\,833.65\,kg \\
Read & Type B segment sheet, quantity table: what is the last row, the total over all segments of the bridge? & 13\,269\,030.66\,kg \\
Read & Type C segment sheet, quantity table: what is the last row, the total over all segments of the bridge? & 596\,085.82\,kg \\
Bind & For the segment type whose welds weigh 4404.58\,kg per segment, how many segments does the bridge have? & 2 \\
\bottomrule
\end{tabular}}
\caption{\textbf{An example of a multi-sheet item in Few-Sheet and Full-Set.} The correct option is the sum of the three values that the read probes ask for, one per evidence page; each decoy carries a quantity table of the same form whose total the question excludes.}
\label{fig:examples-multi}
\end{figure*}

\subsection{Reconstruction and design tasks}
\label{sec:s-rebuild-examples}

\Cref{fig:examples-rebuild} shows the three levels on one set of five approach-slab pages: L1 rebuilds its second page, L2 rebuilds the whole set, and L3 produces its third page from the second and the new slab width.
Inputs are shown as the agent receives them and the ground truth with one colour per layer, so the gap between what is printed and what is scored is visible.

\newcommand{\exrpage}[3]{%
\begin{minipage}[t]{#1\linewidth}\centering
\setlength{\fboxsep}{0pt}\setlength{\fboxrule}{0.3pt}\fbox{#2}\par\vspace{0.25em}
{\scriptsize #3}
\end{minipage}}
\newcommand{\exrchip}[2]{\textcolor[RGB]{#1}{\rule[-0.05ex]{0.9em}{1.1ex}}\hspace{0.3em}#2}
\newcommand{\exrhead}[1]{\begin{minipage}{\linewidth}\raggedright #1\end{minipage}\par\vspace{0.45em}}
\begin{figure*}[p]
\centering\footnotesize
\exrhead{\textbf{(a) L1:} one printed page in, one editable CAD file out. \textit{Task statement:} write \texttt{output.dxf} that restores everything on the page as an editable CAD file, in paper millimetres from the lower-left corner, with every entity on one of nine named layers; dimensions are \texttt{DIMENSION} entities showing the values in the drawing, text is \texttt{TEXT} or \texttt{MTEXT}, hatches are \texttt{HATCH}, hidden and centre lines use their line types, and circles and arcs are true \texttt{CIRCLE} and \texttt{ARC}.}
\exrpage{0.492}{\includegraphics[width=\dimexpr\linewidth-0.6pt\relax]{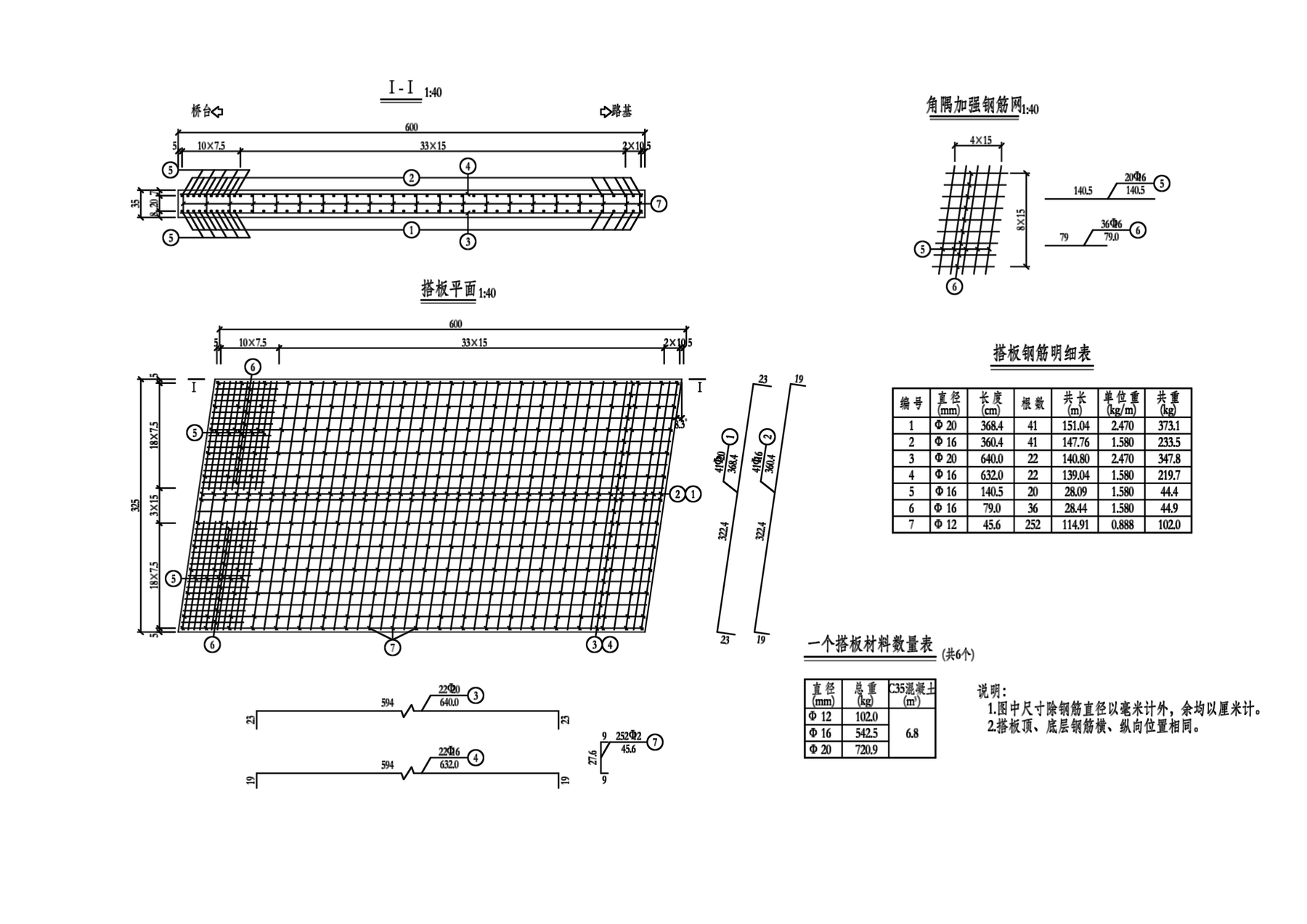}}{input: \texttt{input.pdf}, one raster page}\hfill
\exrpage{0.492}{\includegraphics[width=\dimexpr\linewidth-0.6pt\relax]{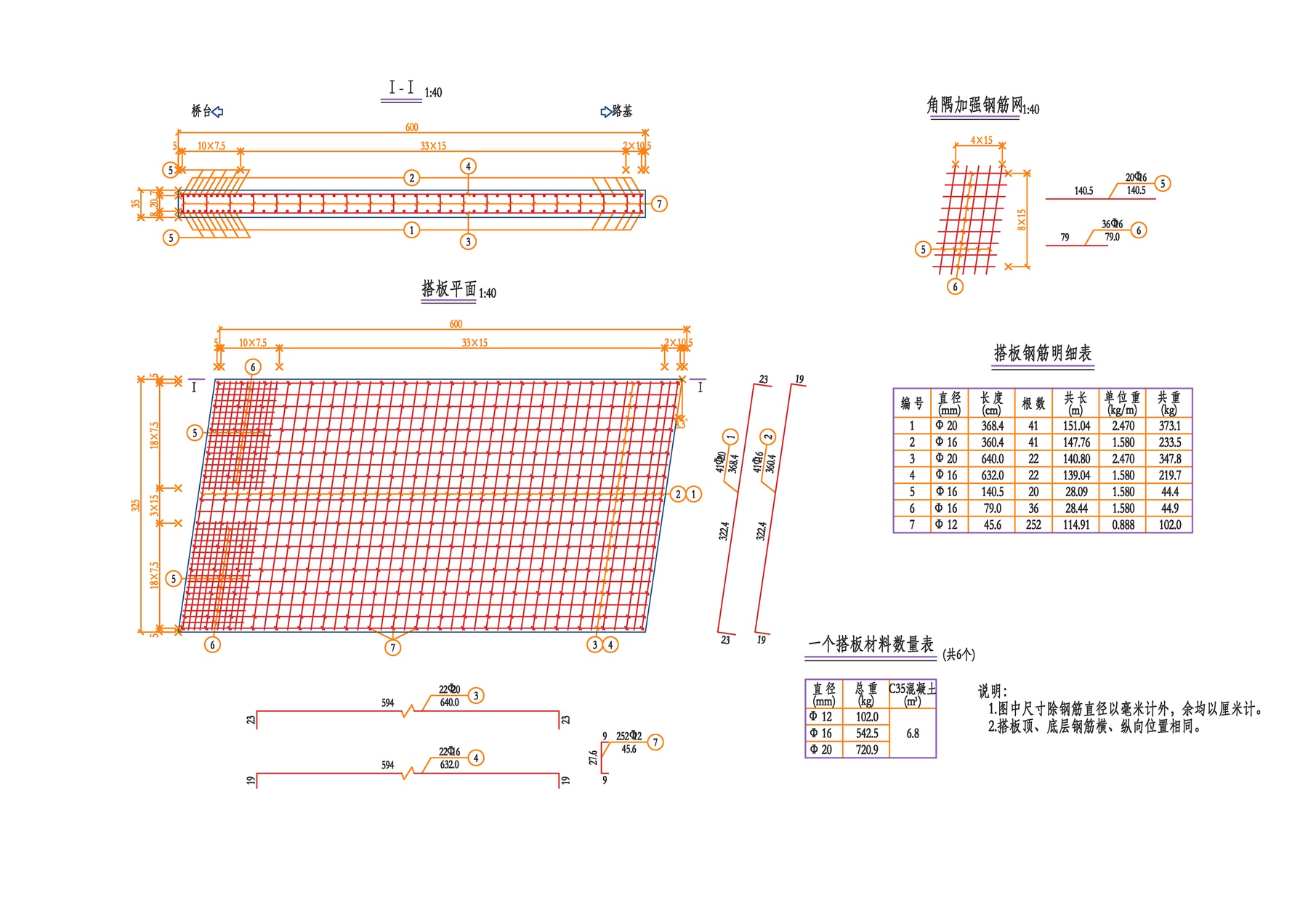}}{ground truth: the designers' file, one colour per layer}

\vspace{0.5em}
{\scriptsize
\exrchip{31,78,140}{outline}\hspace{1.1em}\exrchip{214,39,40}{reinforcement}\hspace{1.1em}\exrchip{120,144,168}{hidden lines}\hspace{1.1em}%
\exrchip{44,160,44}{centre lines}\hspace{1.1em}\exrchip{170,170,170}{hatch}\hspace{1.1em}\exrchip{255,127,14}{dimensions}\hspace{1.1em}%
\exrchip{40,40,40}{text}\hspace{1.1em}\exrchip{148,103,189}{symbols}\hspace{1.1em}\exrchip{140,86,75}{tables}}

\vspace{1.0em}
\exrhead{\textbf{(b) L2:} the five pages of the set in, one file per page out. \textit{Task statement:} as in L1 for every page; this is one set, so keep the layers, text styles and dimension styles the same in every file, and give the same member the same dimensions and bar numbers on every page.}
\exrpage{0.192}{\includegraphics[width=\dimexpr\linewidth-0.6pt\relax]{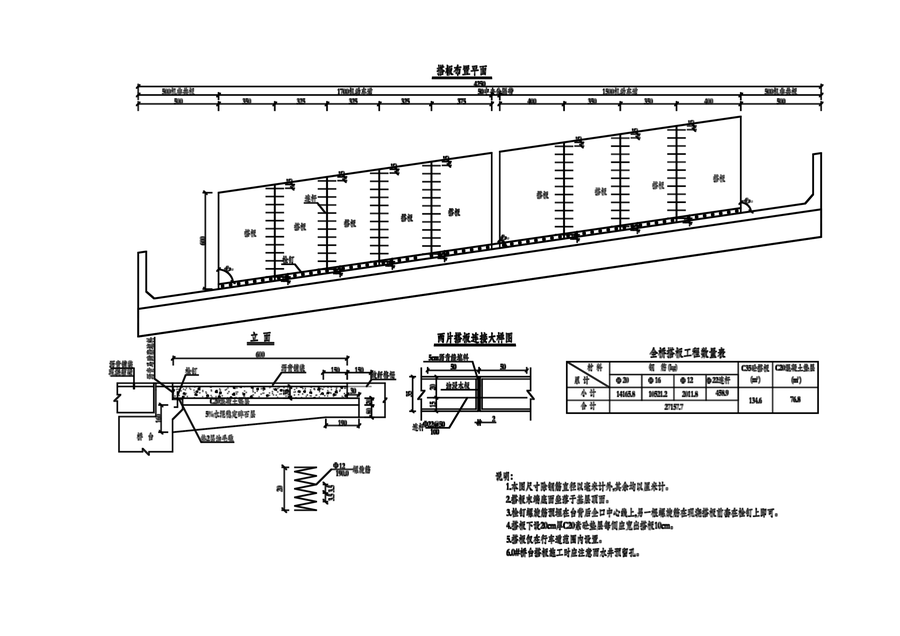}}{general arrangement\\\texttt{page\_01.dxf}}\hfill
\exrpage{0.192}{\includegraphics[width=\dimexpr\linewidth-0.6pt\relax]{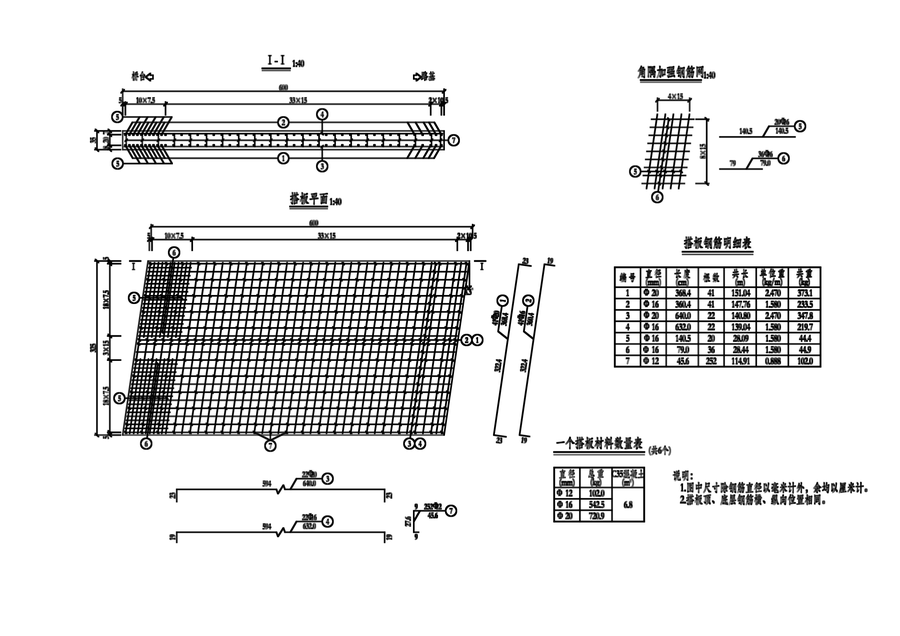}}{reinforcement, page 1 of 4\\\texttt{page\_02.dxf}}\hfill
\exrpage{0.192}{\includegraphics[width=\dimexpr\linewidth-0.6pt\relax]{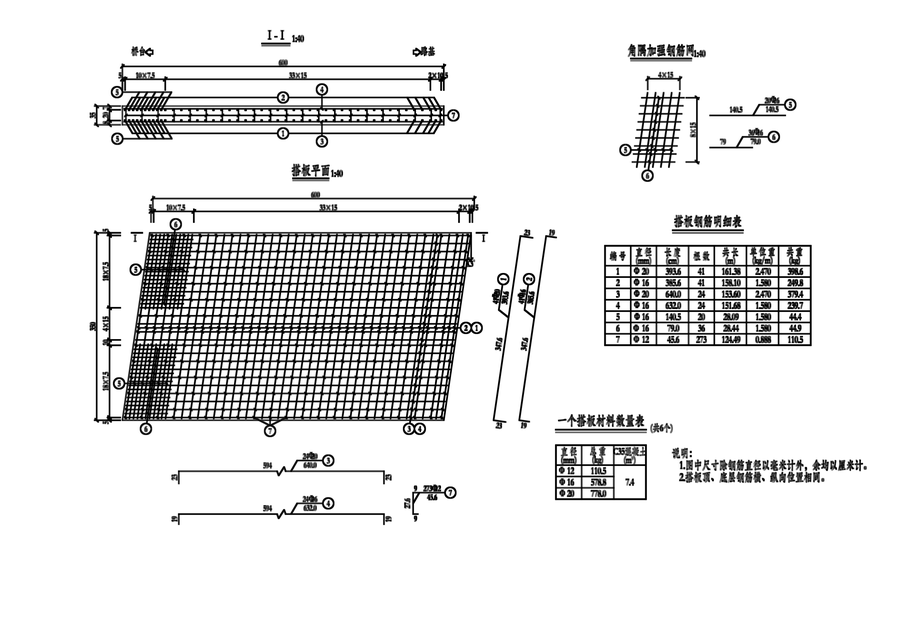}}{reinforcement, page 2 of 4\\\texttt{page\_03.dxf}}\hfill
\exrpage{0.192}{\includegraphics[width=\dimexpr\linewidth-0.6pt\relax]{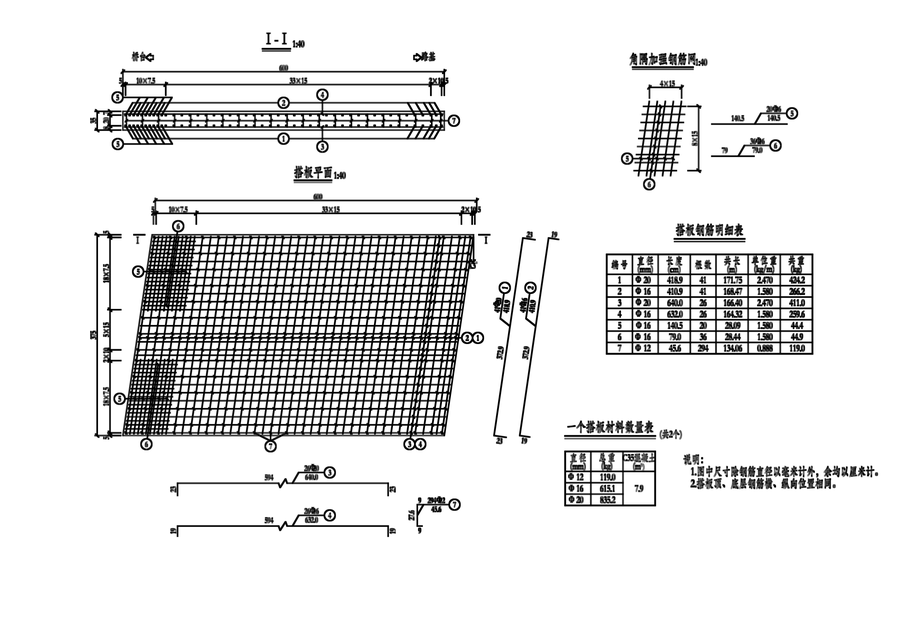}}{reinforcement, page 3 of 4\\\texttt{page\_04.dxf}}\hfill
\exrpage{0.192}{\includegraphics[width=\dimexpr\linewidth-0.6pt\relax]{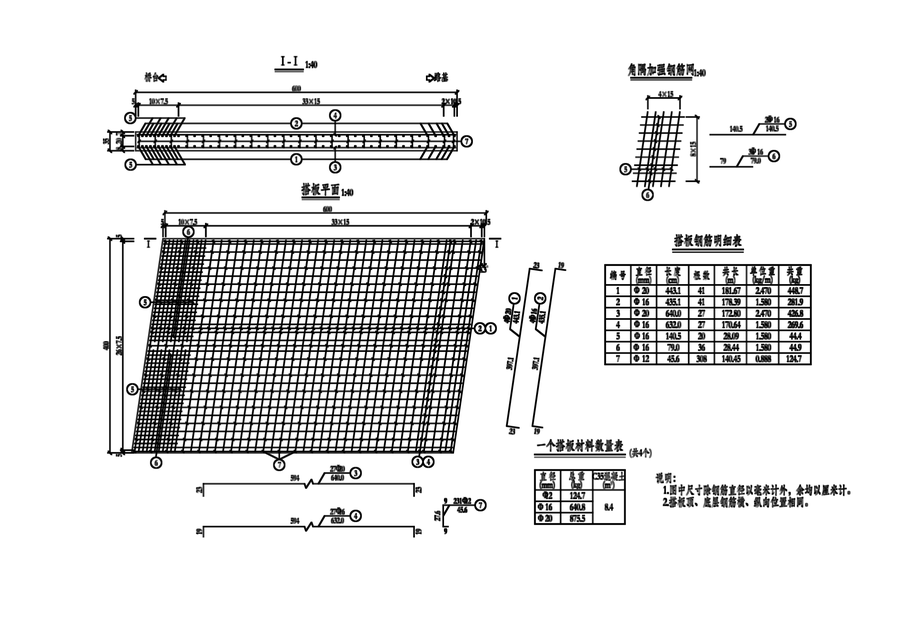}}{reinforcement, page 4 of 4\\\texttt{page\_05.dxf}}

\vspace{1.0em}
\exrhead{\textbf{(c) L3:} the reference page as a CAD file, other pages of the set as rasters and the new parameters in; the withheld page out. \textit{Task statement:} draw page 2 of 4 of the approach slab reinforcement. The slab on this page is 350\,cm wide (325\,cm on the reference page), and there are six slabs of this type; its length, thickness, skew angle, bar diameters and spacing rules are those of the reference page. Recompute the bar counts and layout, the length of each numbered bar and the values of the bar schedule and the material table, and update the dimensions in the drawing.}
\exrpage{0.322}{\includegraphics[width=\dimexpr\linewidth-0.6pt\relax]{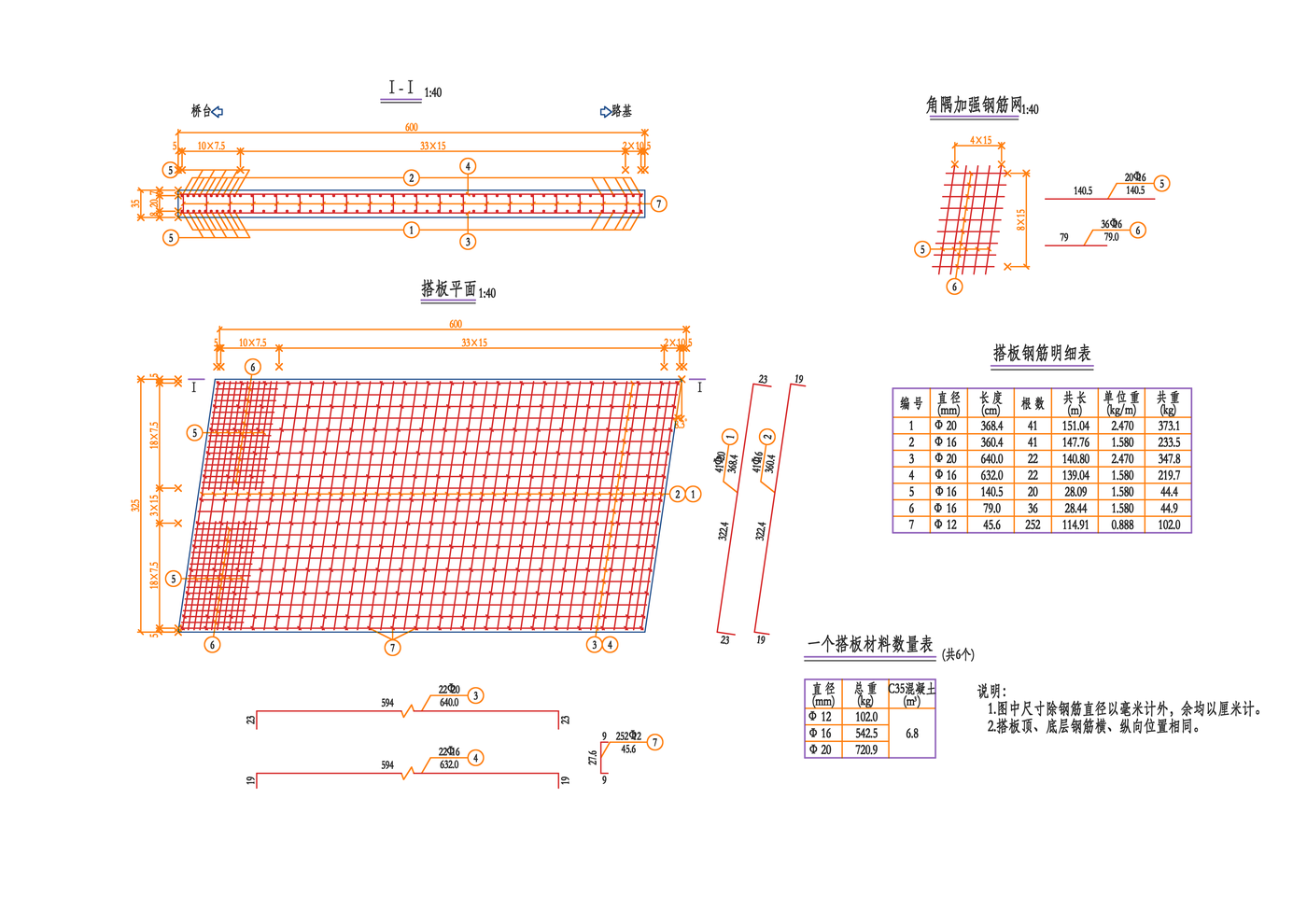}}{reference page: \texttt{ref/sibling.dxf}, 325\,cm}\hfill
\begin{minipage}[t]{0.322\linewidth}\centering
\setlength{\fboxsep}{0pt}\setlength{\fboxrule}{0.3pt}%
\begin{minipage}[b]{\linewidth}
\fbox{\includegraphics[width=\dimexpr0.49\linewidth-0.6pt\relax]{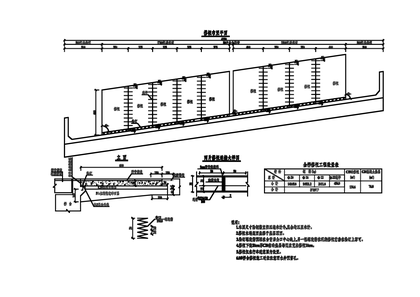}}\hfill
\fbox{\includegraphics[width=\dimexpr0.49\linewidth-0.6pt\relax]{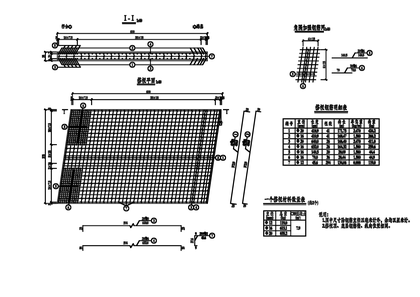}}\par\vspace{0.02\linewidth}
\fbox{\includegraphics[width=\dimexpr0.49\linewidth-0.6pt\relax]{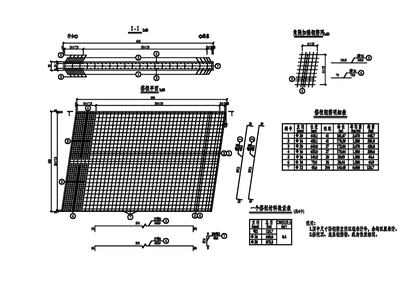}}\hfill
\fbox{\includegraphics[width=\dimexpr0.49\linewidth-0.6pt\relax]{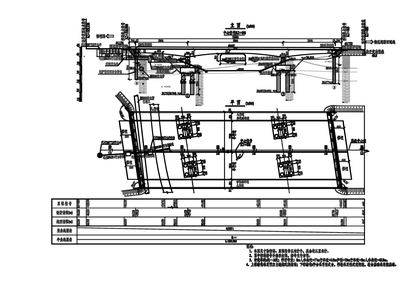}}
\end{minipage}\par\vspace{0.25em}
{\scriptsize other pages: \texttt{ref/context.pdf}}
\end{minipage}\hfill
\exrpage{0.322}{\includegraphics[width=\dimexpr\linewidth-0.6pt\relax]{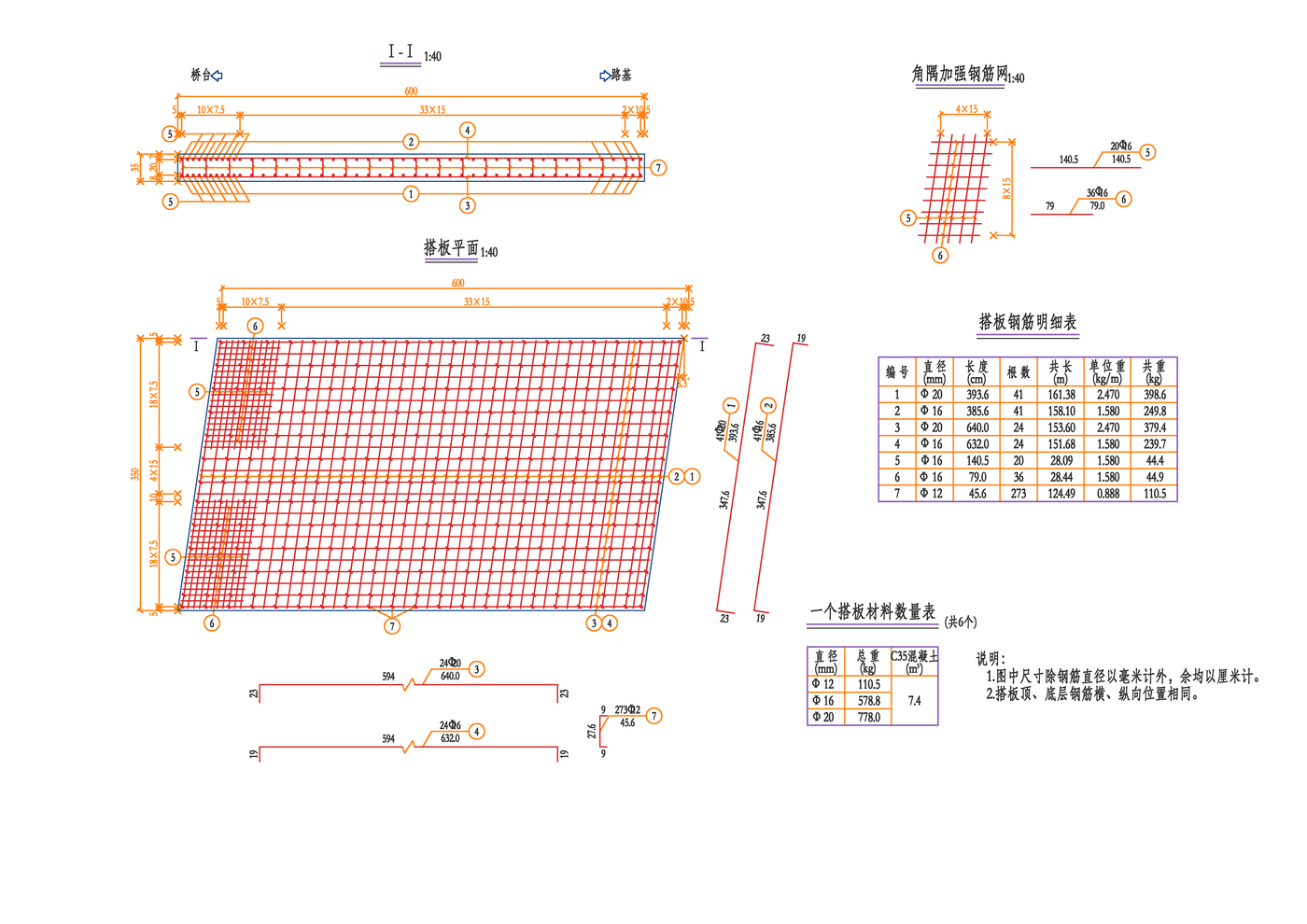}}{withheld page (ground truth), 350\,cm}
\caption{\textbf{Examples of reconstruction and design tasks (L1 to L3) on one set.} The approach-slab set of five pages carries a task at each level: (a) L1 rebuilds its second page, (b) L2 rebuilds the whole set and (c) L3 produces its third page from the second and the new slab width. Inputs are shown as the agent receives them, and CAD files are plotted with one colour per layer.}
\label{fig:examples-rebuild}
\end{figure*}

\subsection{One reconstruction episode}
\label{sec:s-episode}

\begin{figure*}[t]
\centering
\includegraphics[width=\linewidth]{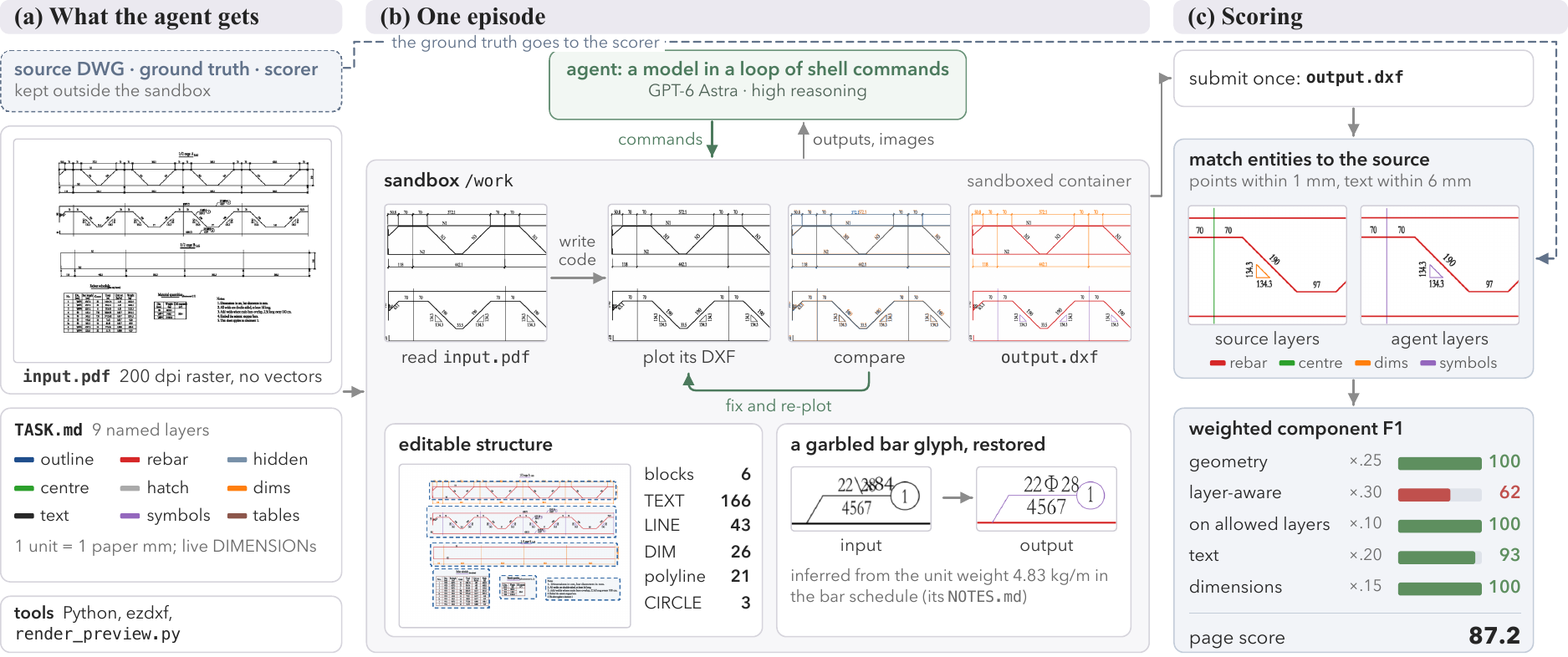}
\caption{\textbf{An example of an L1 episode: input, agent loop and scoring.}
(a) The agent receives one raster page, a task card that fixes paper-millimetre coordinates and nine named layers, and tools; the source file, the ground truth and the scorer stay outside the sandbox.
(b) It writes and runs its own code, plots its DXF in the input's style and compares the plot with the input (grey: in both; blue: input only; orange: output only), then submits once. Its file groups the page into six named blocks, and it restores a garbled bar glyph of the input from the unit weight in the bar schedule.
(c) The scorer matches entities with the source file and weights five component F1 scores; here the agent puts a centre line and a slope triangle on the symbol layer, where the source has them on its centre-line and dimension layers.
L2 runs the same loop over a whole set, and L3 starts from a reference page and new parameters.}
\label{fig:episode}
\end{figure*}

\Cref{fig:episode} follows one L1 run from the input to the score.
The agent works in a sandbox: it receives a raster page, a task card that fixes paper-millimetre coordinates and nine named layers, and tools, while the source file, the ground truth and the scorer stay outside.
It writes and runs its own code, plots its DXF in the input's style, compares the plot with the input and submits once.
In this run the agent groups the page into six named blocks and restores a garbled bar glyph of the input from the unit weight in the bar schedule, which shows the reasoning that L1 asks for beyond copying lines.
The scorer then matches entities with the source file and weights five component F1 scores; here the agent puts a centre line and a slope triangle on the symbol layer, where the source has them on its centre-line and dimension layers, a small instance of the layer errors discussed in \cref{sec:analysis}.

\subsection{Case studies}
\label{sec:s-cases}

\begin{figure*}[p]
\centering\footnotesize
\exrhead{\textbf{(a) One page, alone and within its set.} GPT-6~Astra rebuilds an abutment page alone and as page 2 of a 13-page set. \mbox{\exrchip{255,127,14}{dimension entities}}; all else in grey.}
\exrpage{0.325}{\includegraphics[width=\dimexpr\linewidth-0.6pt\relax]{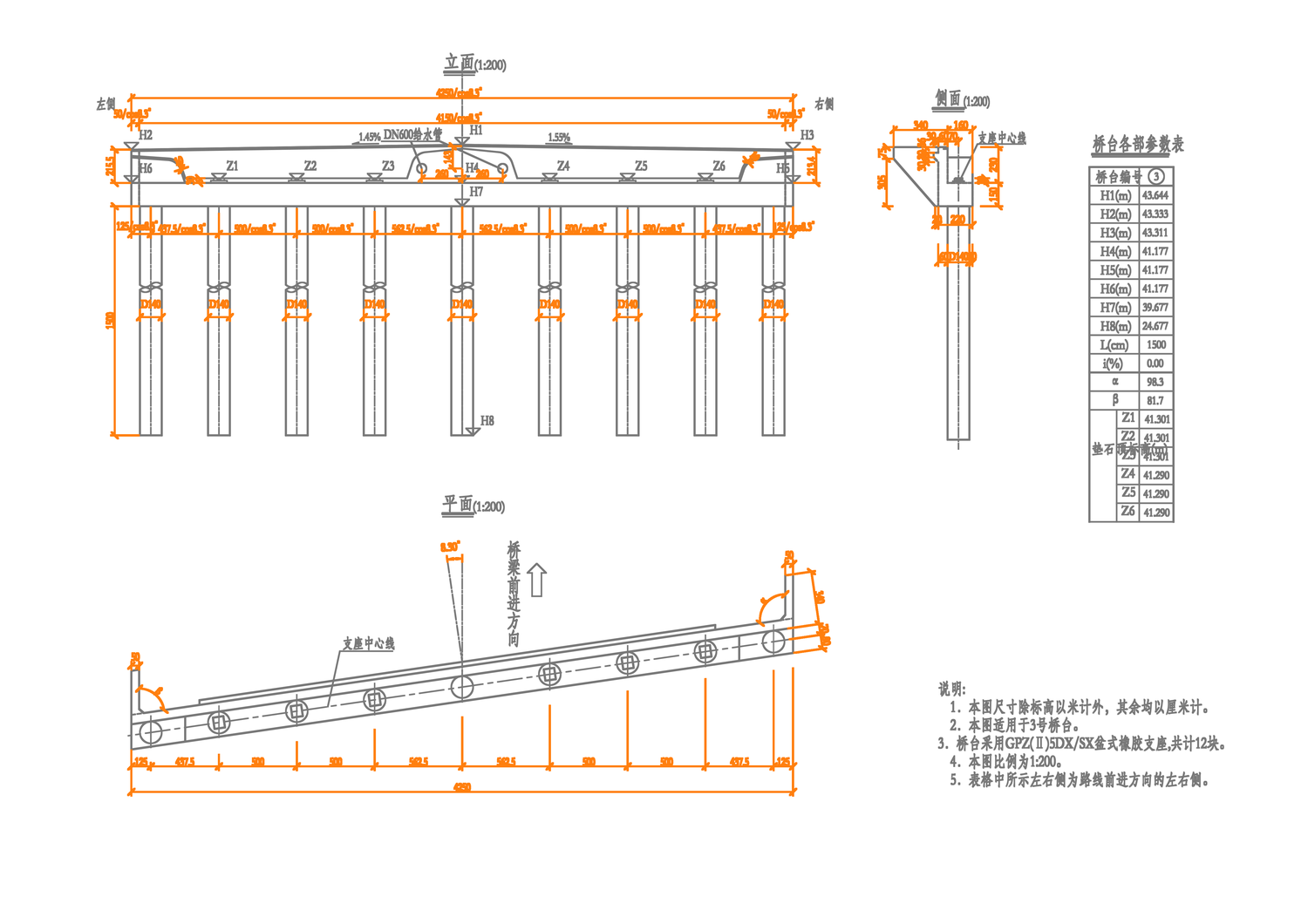}}{designer's file: 69 dimension entities}\hfill
\exrpage{0.325}{\includegraphics[width=\dimexpr\linewidth-0.6pt\relax]{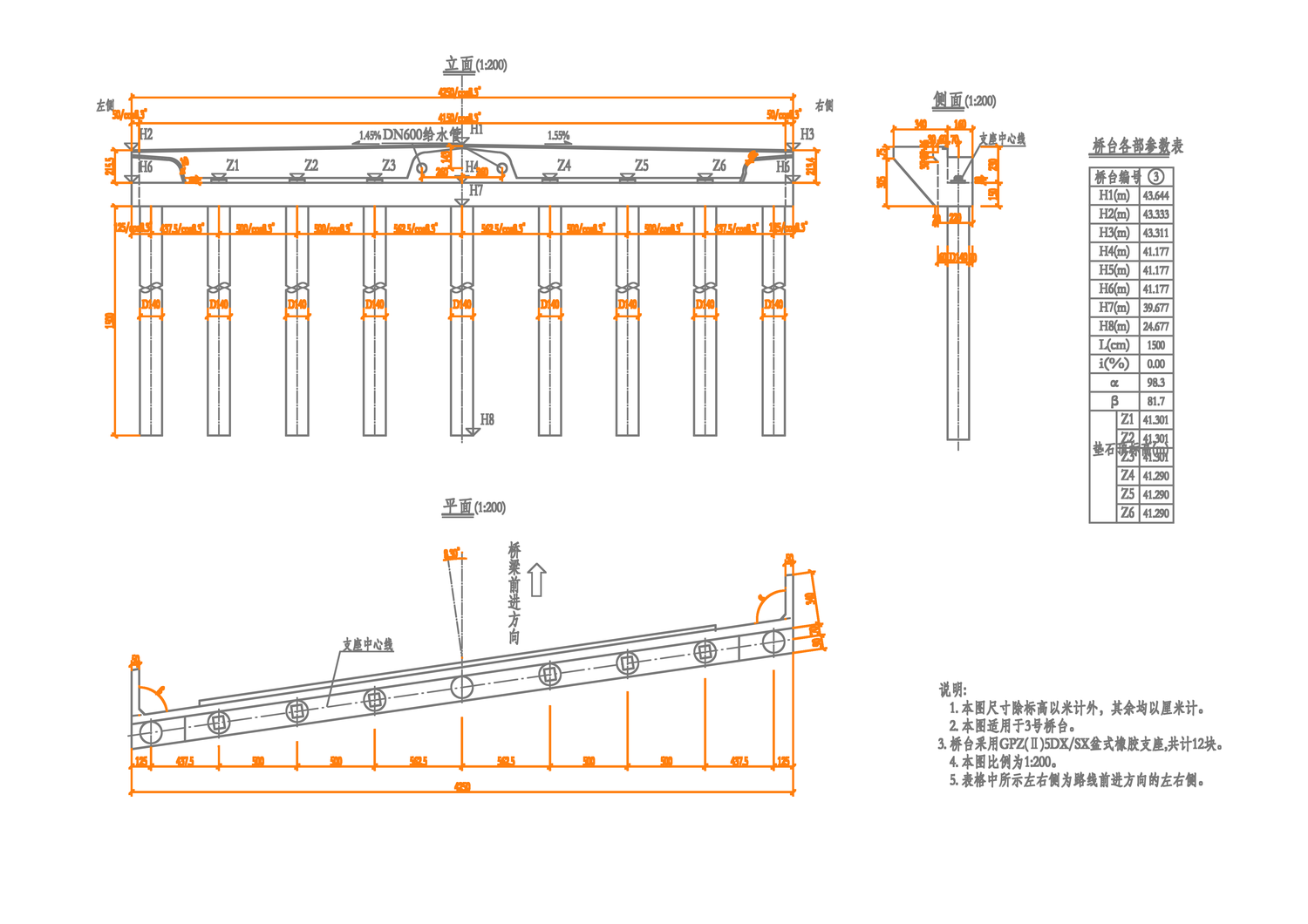}}{alone: score 98.2, 69 dimension entities\\geometry 99.6, text 98.7, dimension 100}\hfill
\exrpage{0.325}{\includegraphics[width=\dimexpr\linewidth-0.6pt\relax]{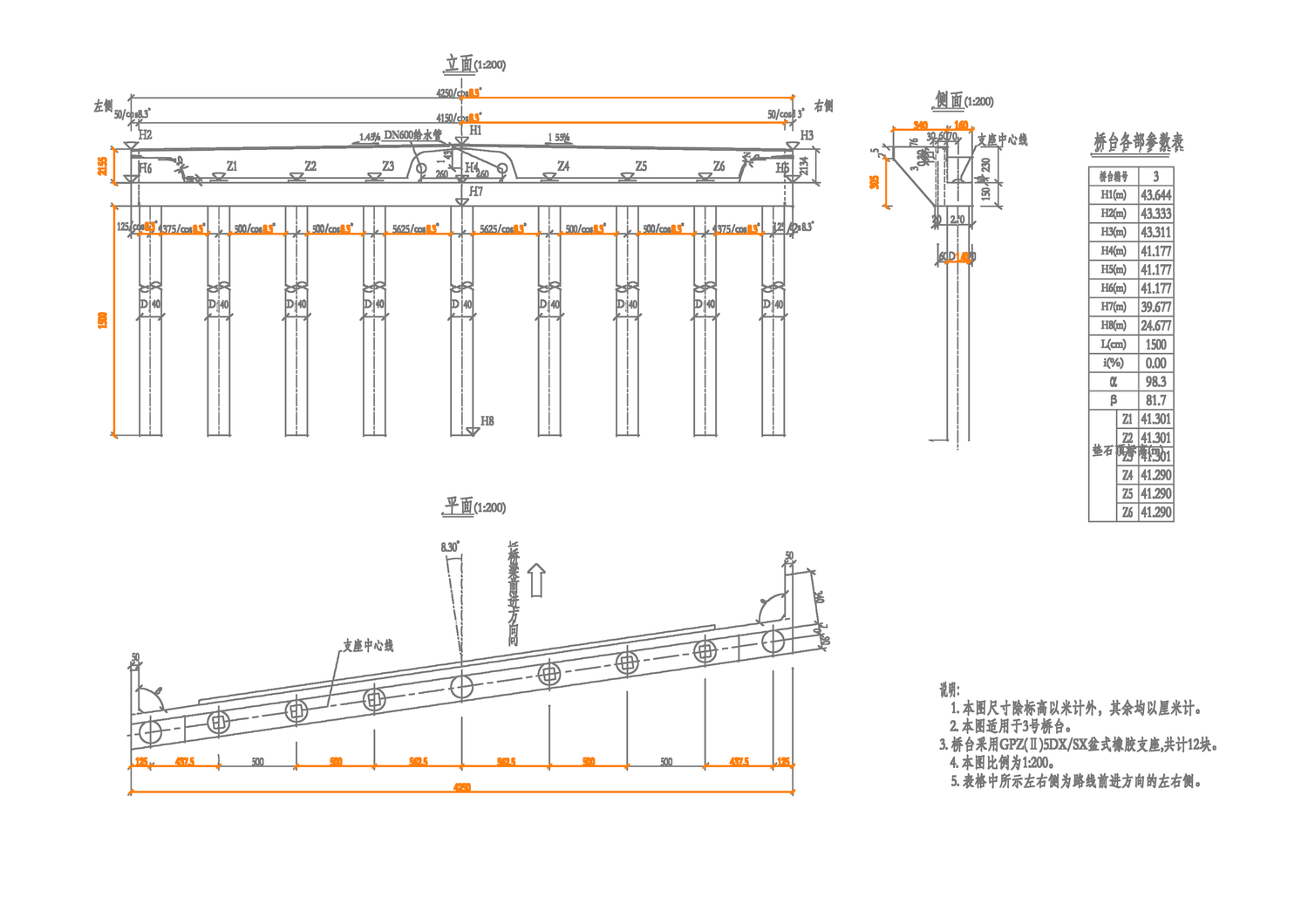}}{within the set: score 76.7, 26 dimension entities\\geometry 97.5, text 70.2, dimension 36.0}

\vspace{1.2em}
\exrhead{\textbf{(b) One page, four models.} Single-page reconstruction of a box-girder section with its reinforcement.}
\exrpage{0.192}{\includegraphics[width=\dimexpr\linewidth-0.6pt\relax]{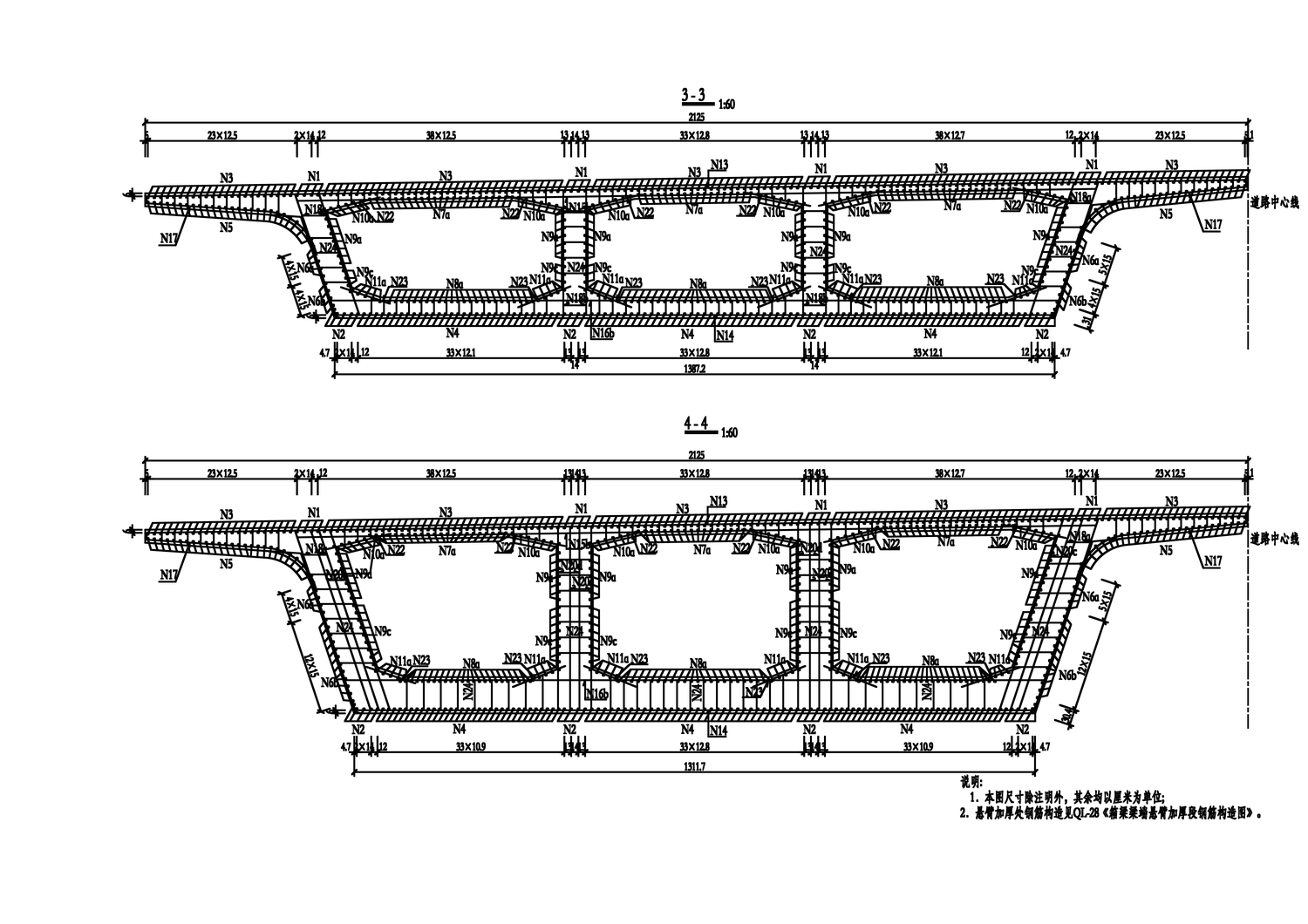}}{designer's file\\\textit{page score}\\\textit{geometry F1}\\\textit{text F1}\\\textit{dimension F1}}\hfill
\exrpage{0.192}{\includegraphics[width=\dimexpr\linewidth-0.6pt\relax]{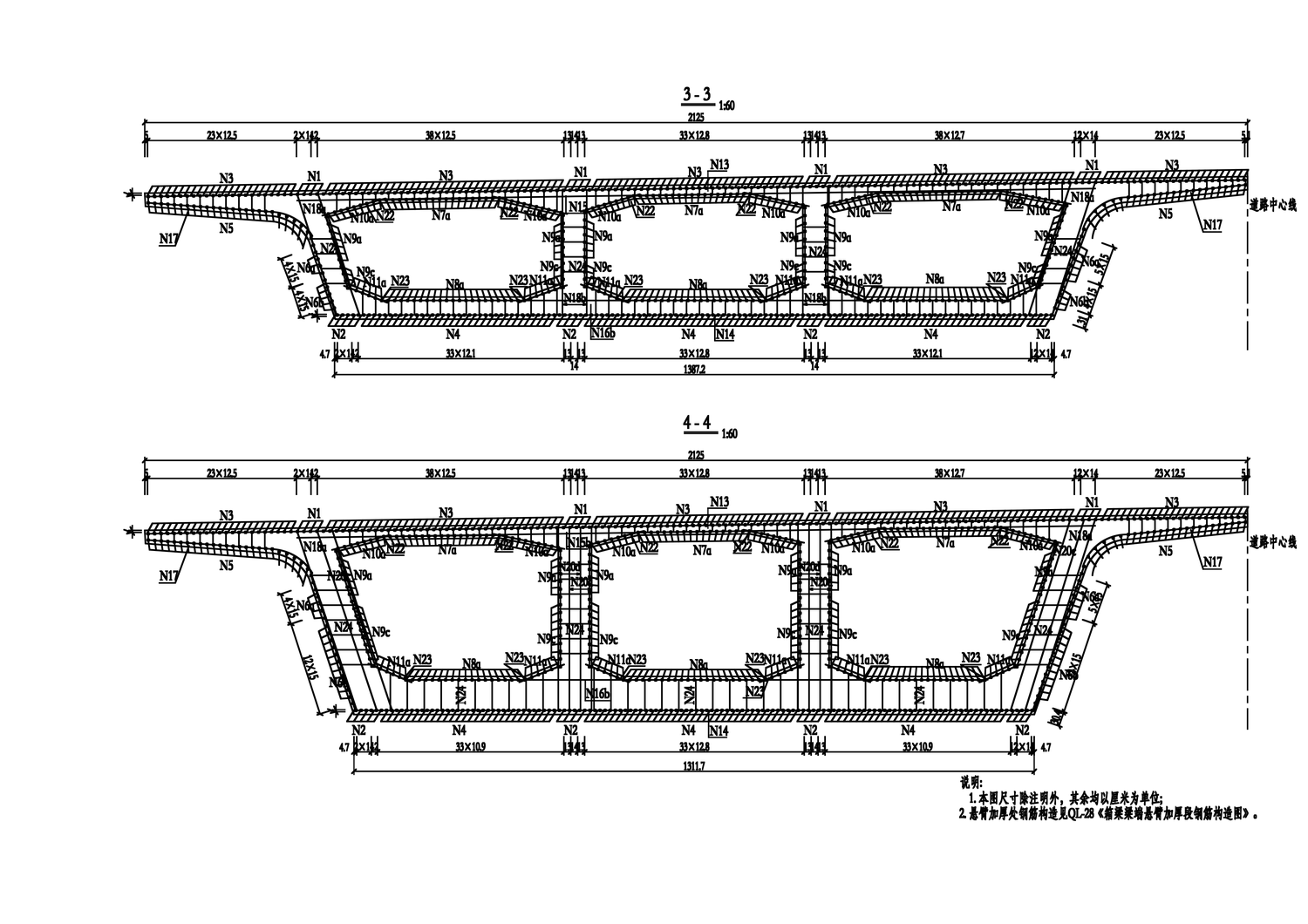}}{GPT-6~Astra\\97.7\\97.9\\99.1\\93.8}\hfill
\exrpage{0.192}{\includegraphics[width=\dimexpr\linewidth-0.6pt\relax]{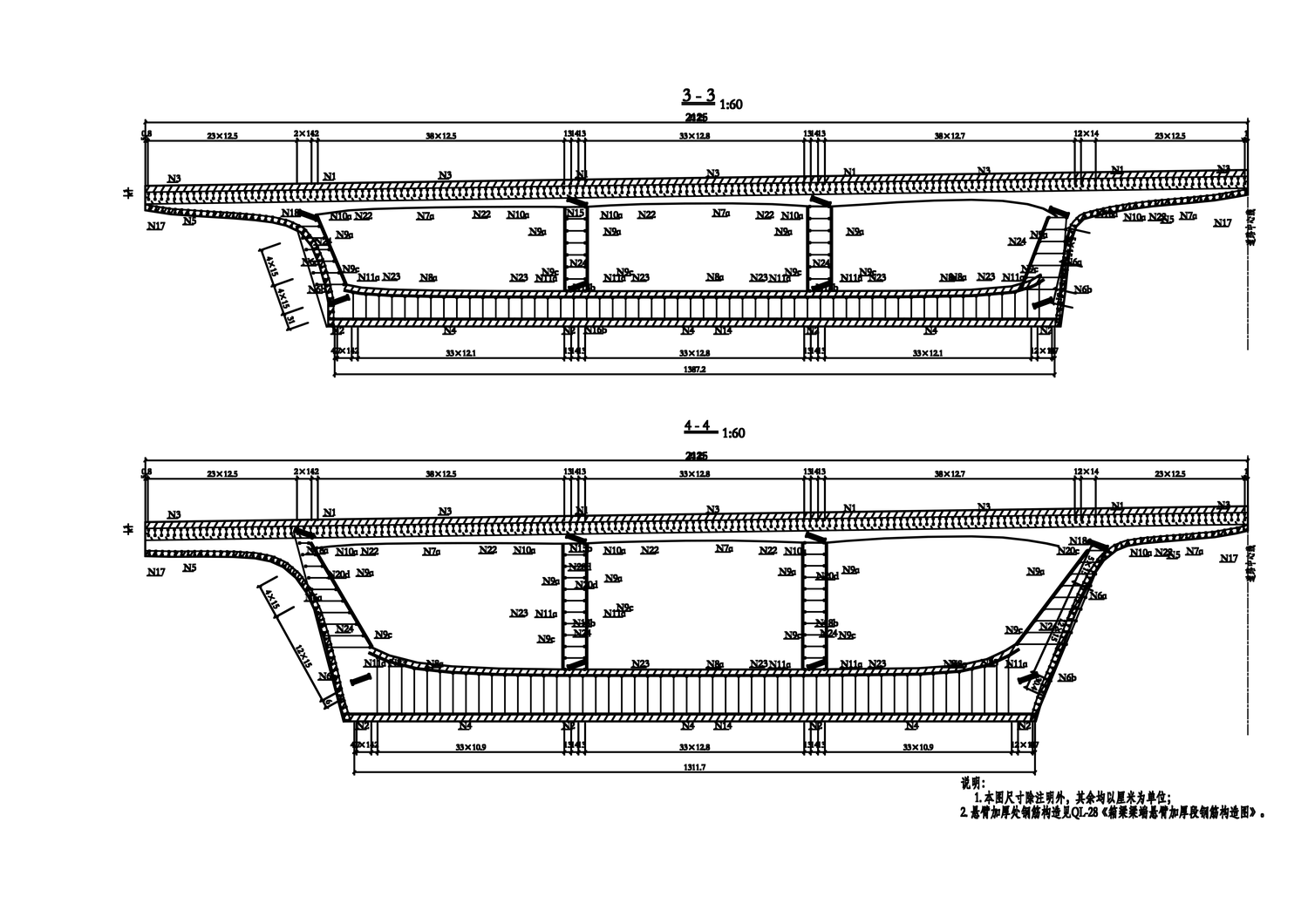}}{DeepSeek-V4.1-Flash\\81.2\\84.0\\65.1\\85.5}\hfill
\exrpage{0.192}{\includegraphics[width=\dimexpr\linewidth-0.6pt\relax]{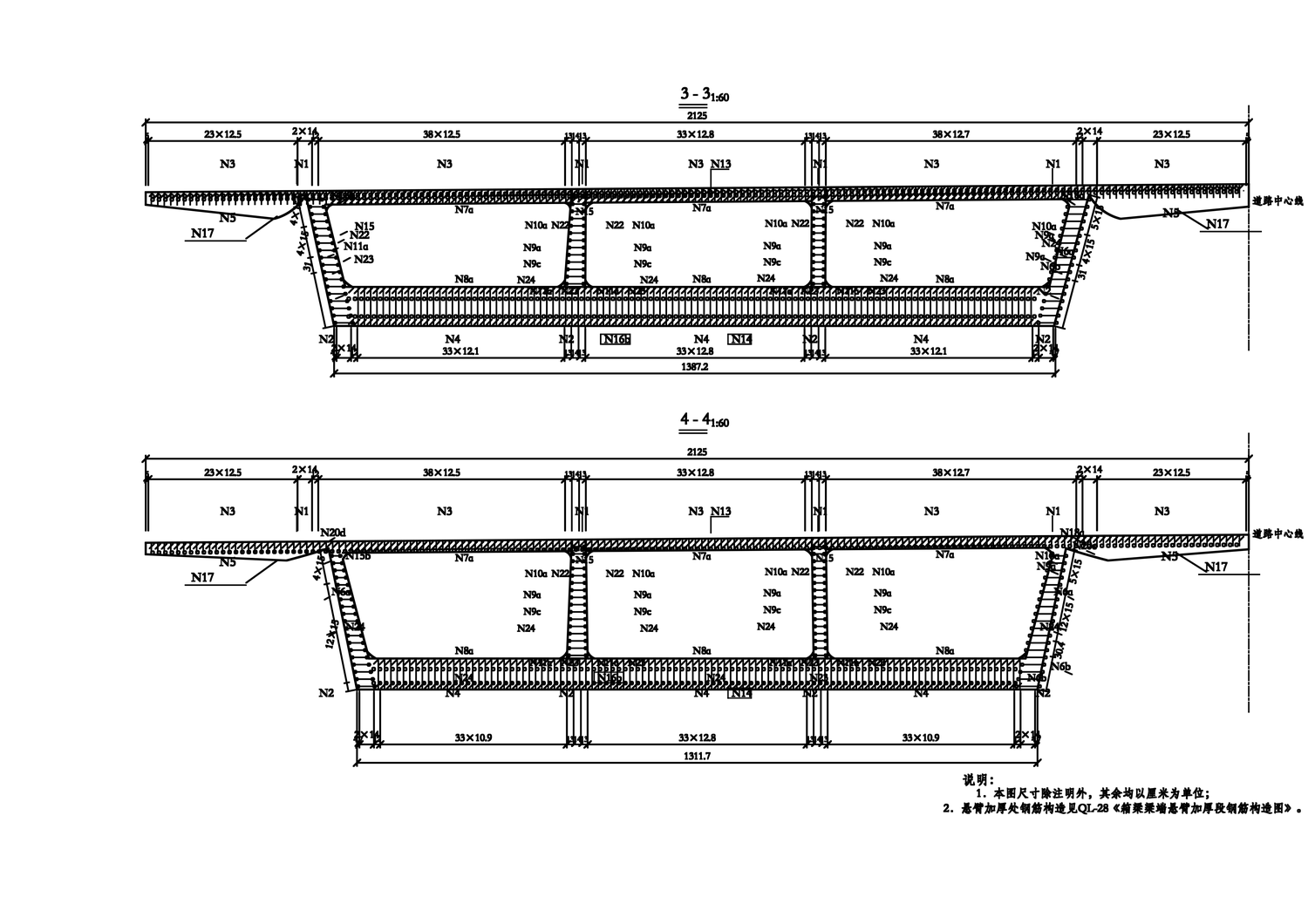}}{GLM-5.3-Flash\\72.2\\70.2\\47.9\\89.5}\hfill
\exrpage{0.192}{\includegraphics[width=\dimexpr\linewidth-0.6pt\relax]{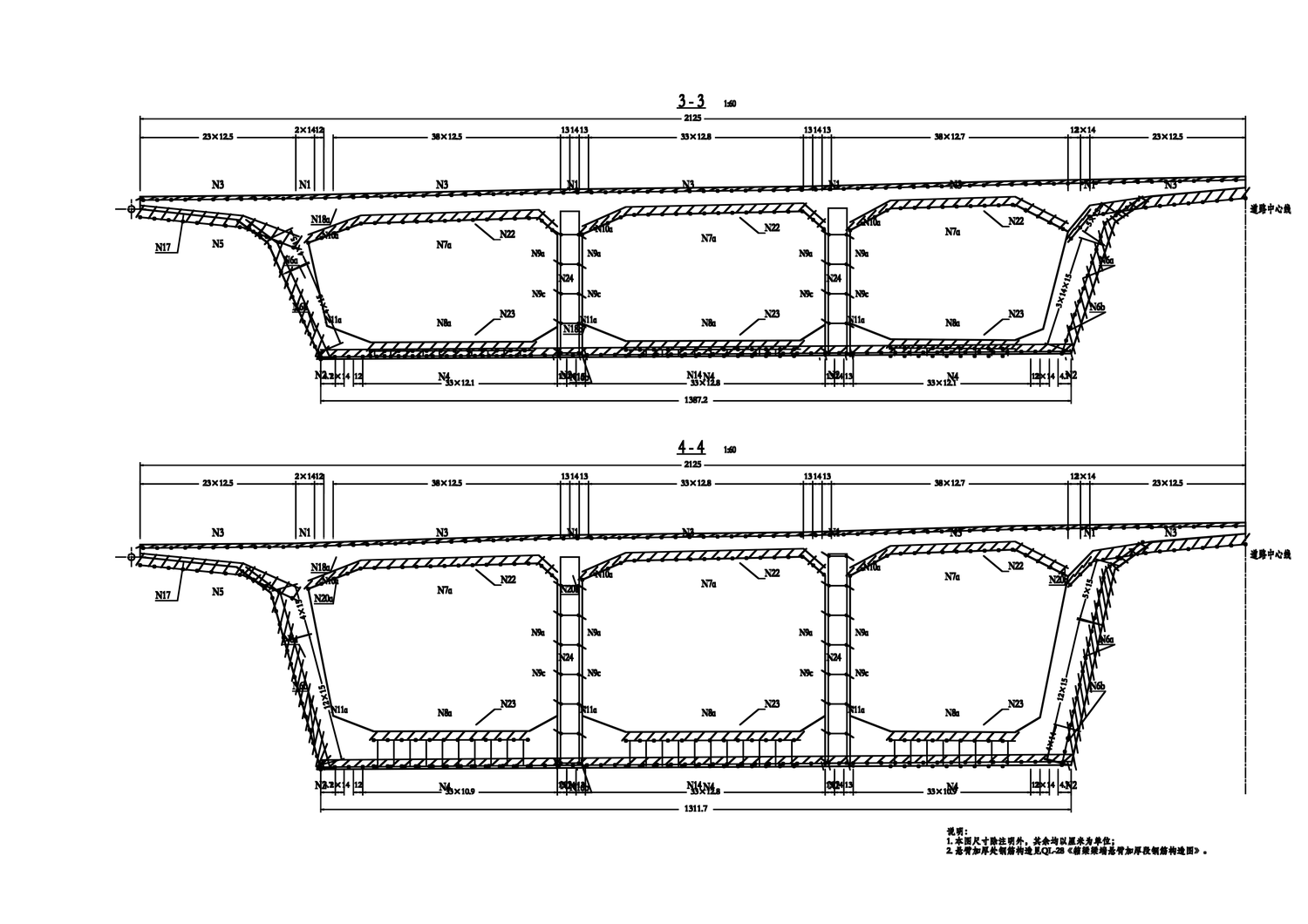}}{GPT-5.6~Sol\\46.4\\46.7\\22.5\\42.0}

\vspace{1.2em}
\exrhead{\textbf{(c) A count across a sheet seam.} \textit{Parent question:} The central reservation of a road in an interchange area is planted with alternating groups of purple-leaf plum and yellow trumpet trees, over alternating segments of golden privet and ruellia ground cover. From the west end of the reservation to the east end, how many trees stand on the golden-privet segments?\quad
\textbf{A}~24~\cmark\quad \textbf{B}~28\quad \textbf{C}~16\quad \textbf{D}~12\quad \textbf{E}~66\quad \textbf{F}~42}
\exrhead{\textit{Evidence:} the reservation runs across two tree plans (pages 17 and 20 of the 38-page unit) and two ground-cover plans (pages 32 and 35). The plum groups, four trees each, stand on the golden-privet segments. Page 17 labels three groups; page 20 draws four, the first of which is the last group of page 17 drawn again without a label. The count is $(3+4-1)\times 4=24$. Option B counts the group at the seam twice, $(3+4)\times 4=28$, and option C counts the groups of page 20 only. One probe asks which group is drawn on both plans.}
\begin{tabular}{@{}l@{\hspace{2.2em}}cc@{\hspace{2.2em}}cccc@{}}
\toprule
 & \multicolumn{2}{c@{\hspace{2.2em}}}{Few-Sheet} & \multicolumn{4}{c}{Full-Set} \\
\cmidrule(r{2.2em}){2-3}\cmidrule{4-7}
Model & Answer & Probes right & Answer & Probes right & Pages opened & Evidence pages opened \\
\midrule
GPT-6~Astra & A (24)~\cmark & 6 of 6 & B (28) & 6 of 6 & 18 & 4 of 4 \\
GPT-6.1~Sol & A (24)~\cmark & 6 of 6 & B (28) & 6 of 6 & 20 & 4 of 4 \\
GPT-5.6~Sol & A (24)~\cmark & 6 of 6 & C (16) & 5 of 6 & 20 & 3 of 4 \\
DeepSeek-V4.1-Flash & A (24)~\cmark & 6 of 6 & D (12) & 4 of 6 & 20 & 4 of 4 \\
Qwen3.8-27B & C (16) & 6 of 6 & B (28) & 4 of 6 & 17 & 1 of 4 \\
GLM-5.3-Flash & B (28) & 6 of 6 & B (28) & 4 of 6 & 18 & 2 of 4 \\
Qwen3.5-9B & B (28) & 1 of 6 & E (66) & 0 of 6 & 10 & 0 of 4 \\
\bottomrule
\end{tabular}
\caption{\textbf{Case studies of reconstruction (a, b) and understanding (c).} (a) Within the set, the agent keeps 26 of the 69 dimensions as dimension entities and redraws the rest as loose lines whose labels are truncated or missing: the dimension score falls from 100 to 36.0 and the stray text lowers the text score. (b) The outputs range from a page that matches the designer's file to one that keeps the section outline and loses most of the reinforcement and its labels. (c) In Full-Set, GPT-6~Astra and GPT-6.1~Sol open all four evidence pages and answer every probe correctly, including the one on the seam group, yet count that group twice.}
\label{fig:cases}
\end{figure*}

\Cref{fig:cases} follows three results of \cref{tab:main} down to single items.
Panel (a) shows the set penalty of \cref{sec:exp-rebuild}: the abutment page that GPT-6 Astra rebuilds with a score of 98.2 when alone scores 76.7 as page 2 of a 13-page set, because it keeps 26 of the 69 dimensions as dimension entities and redraws the rest as loose lines with truncated or missing labels.
Panel (b) shows how far the outputs on one page spread, from a page that matches the designer's file to one that keeps the section outline and loses most of the reinforcement and its labels.
Panel (c) follows a Full-Set question in which GPT-6 Astra and GPT-6.1 Sol open all four evidence pages and answer every probe correctly, including the one on the seam group, and yet count that group twice, a failure of combining and not of reading.

\end{document}